%% file: acl_latex.tex
\pdfoutput=1
\documentclass[11pt]{article}

\usepackage[final]{acl}

\usepackage{times}
\usepackage{latexsym}
\usepackage{enumitem}
\usepackage{amsmath}

\usepackage[T1]{fontenc}
\usepackage[utf8]{inputenc}

\usepackage{microtype}

\usepackage{inconsolata}

\usepackage{graphicx}
\usepackage{tcolorbox}

\usepackage{booktabs}
\usepackage{tabularx}
\usepackage{multirow}
\usepackage{multicol}
\usepackage{bm}
\newcommand{\bl}[1]{\textcolor{black}{#1}}

\newtcolorbox{sharp_box}{
    sharpish corners, % better drop shadow
    boxrule = 0pt,
    toprule = 4.5pt, % top rule weight
}

\title{Stress Testing Factual Consistency Metrics for Long-Document Summarization}

\author{Zain Muhammad Mujahid \and Dustin Wright \and Isabelle Augenstein \\ 
    University of Copenhagen\\
    \texttt{\{zamu, dw, augenstein\}@di.ku.dk}}

\begin{document}
\maketitle
\begin{abstract}
Evaluating the factual consistency of abstractive text summarization remains a significant challenge, particularly for long documents, where conventional metrics struggle with input length limitations and long-range dependencies.
% In this work, we systematically evaluate the reliability of factuality metrics under a range of controlled perturbations, including shuffling source segments, injecting source tags, negating relations, and applying lexical paraphrases that preserve semantic alignment.
In this work, we systematically evaluate the reliability of six widely used reference-free factuality metrics, originally proposed for short-form summarization, in the long-document setting.
% (using a sentence-level retrieval-based scoring strategy that enables evaluation on arbitrarily long inputs?)
% To accommodate extended contexts, we adopt the LongDocFactScore framework, which decomposes summaries into individual sentences, retrieves the most relevant source snippets for each sentence, and aggregates sentence‑level scores into a summary score.
% Through extensive experiments in both single‑ and multi‑document settings, we expose significant weaknesses in existing metrics.
We probe metric robustness through seven factuality-preserving perturbations applied to summaries, namely paraphrasing, simplification, synonym replacement, logically equivalent negations, vocabulary reduction, compression, and source text insertion, and further analyze their sensitivity to retrieval context and claim information density.
Across three long-form benchmark datasets spanning science fiction, legal, and scientific domains, our results reveal that existing short-form metrics produce inconsistent scores for semantically equivalent summaries and exhibit declining reliability for information-dense claims whose content is semantically similar to many parts of the source document.
While expanding the retrieval context improves stability in some domains, no metric consistently maintains factual alignment under long-context conditions.
Finally, our results highlight concrete directions for improving factuality evaluation, including multi-span reasoning, context-aware calibration, and training on meaning-preserving variations to enhance robustness in long-form summarization.\footnote{We release all code, perturbed data, and scripts required to reproduce our results at \url{https://github.com/zainmujahid/metricEval-longSum}.}
% \footnote{We release our experimental code and all perturbed summary outputs to facilitate future research.}
\end{abstract}

\section{Introduction}
\input{figures/head_figure}

Abstractive summarization has seen rapid advances with the advent of large language models (LLMs), but ensuring that generated summaries faithfully reflect the source content remains a persistent challenge~\cite{DBLP:conf/emnlp/LabanFXW24,DBLP:journals/corr/abs-2502-14409}.
Summaries that read fluently can nonetheless introduce hallucinated details or omit critical facts~\cite{belem-etal-2025-single}, undermining their reliability for downstream tasks in domains such as medicine, law, and scientific review~\cite{DBLP:journals/corr/abs-2411-14199}.
Traditional evaluation measures like ROUGE~\cite{lin2004rouge} and BLEU~\cite{DBLP:conf/acl/PapineniRWZ02}, which rely on n‑gram overlap with reference summaries, are useful for measuring surface similarity but fail to capture factual consistency,
% that is, whether the information in the summary is accurate with respect to the source document,
since two summaries can overlap heavily in wording while still differ in correctness~\cite{maynez-etal-2020-faithfulness}.
This gap has motivated the development of reference‑free factuality evaluation metrics, which assess whether the statements in a summary are supported by the source document itself rather than by comparison with a human-written reference, using techniques such as question answering~\cite{wang-etal-2020-asking, scialom-etal-2021-questeval, fabbri-etal-2022-qafacteval}, natural language inference~\cite{DBLP:journals/tacl/LabanSBH22, chen-eger-2023-menli, DBLP:conf/acl/ZhaYLH23}, or LLM‑based scoring~\cite{DBLP:conf/emnlp/LiuIXWXZ23, fu-etal-2024-gptscore}.

While such metrics have demonstrated promise on short‑document datasets, their scalability to long‑form summarization is likely to be hindered by challenges unique to long context lengths~\cite{DBLP:journals/corr/abs-2408-14906,DBLP:conf/iclr/SarthiATKGM24,DBLP:journals/corr/abs-2404-16130}. %papers about challenges
Important details may be dispersed across hundreds or thousands of tokens and thus overlooked by metrics that process only truncated inputs~\cite{DBLP:conf/emnlp/LabanFXW24}; multi‑document summaries must reconcile diverse writing styles and potentially conflicting information~\cite{DBLP:journals/corr/abs-2411-14199}; and the reference‑free setting deprives evaluators of gold annotations, necessitating robust intrinsic evaluation protocols.
Our goal is to systematically benchmark widely used factual consistency metrics under these long-document conditions, in order to reveal their robustness and limitations.

% To do so, we focus on six widely-used reference-free metrics: BARTScore~\cite{DBLP:conf/nips/YuanNL21}, SummaC-Conv and SummaC-ZS~\cite{DBLP:journals/tacl/LabanSBH22}, AlignScore~\cite{DBLP:conf/acl/ZhaYLH23}, UniEval~\cite{DBLP:conf/emnlp/Zhong0YMJLZJH22}, and MiniCheck~\cite{DBLP:conf/emnlp/TangLD24}.
% While some of these metrics have been benchmarked widely on short-context datasets~\cite{ramprasad2024automaticfactualitymetricsmeasure}, a systematic evaluation which uncovers the unique pitfalls of these metrics on long-context source texts has not been performed.
% To accomplish this, we follow previous work~\cite{DBLP:conf/coling/BishopAX24} and employ a sentence-level retrieval-based scoring strategy (\S~\ref{S:LDFact}), which overcomes token-limit constraints by segmenting summaries into sentences and retrieving the most relevant source snippets for evaluation. This setup allows us to assess the robustness of the above metrics in long-form and multi-document settings.

Building on this foundation, we apply a stress-testing methodology established in previous work~\cite{ramprasad2024automaticfactualitymetricsmeasure} to six widely-used reference-free metrics (\S~\ref{S:ExpSetup}) across seven factuality-preserving perturbations reflecting realistic long‑form summarization phenomena.
% These perturbations broadly cover controlled reorderings of summary segments, targeted insertion of source document tags, entity‑level flips, and paraphrasing operations.
These perturbations (\S~\ref{S:Perturbations}) broadly cover paraphrasing operations, simplification of complex constructions, synonym replacement, reduced lexical diversity, logically equivalent negations, further compression of the summary, and insertion of unrelated source sentences.
They are designed to challenge metrics to distinguish genuine factual consistency from superficial cues and to maintain robustness in the face of lexical and structural variation. On top of this, we investigate the impact of long-document specific phenomena, including retrieval length and evidence dispersion~\cite{DBLP:conf/emnlp/GoldmanJSMDT24}.

Through extensive experiments on six evaluation metrics across three benchmark datasets in science fiction, legal, and scientific domains, we expose significant weaknesses in existing metrics, such as inconsistent scoring across semantically equivalent summaries (Fig.~\ref{Results:fig1}).
Our analysis reveals that current factuality metrics vary widely in robustness to meaning-preserving edits, with some highly sensitive to surface changes while others remain more stable.
Most metrics benefit from broader retrieval context windows, though with notable domain-specific variation.
We also find that metric reliability decreases for information-dense claims that overlap semantically with large portions of the source document, suggesting that current metrics struggle with compressed or globally entangled content.
These insights point toward improving evaluation consistency by developing metrics that integrate multi-span reasoning and context-aware calibration.

% The following summarizes our key contributions:

% \begin{itemize}[noitemsep]
%     \item We conduct a systematic robustness evaluation of six factuality metrics across seven factuality-preserving perturbations on three long-document summarization datasets that span science-fiction, legal, and scientific domains.
%     \item We provide a comprehensive analysis of metric-specific vulnerabilities, revealing that surface-level changes significantly impact scores despite preserving factual consistency.
%     \item We further investigate how retrieval context and claim information density affect metric behaviour, showing that both factors influence evaluation stability across domains.
%     %\item We release our experimental code and all perturbed summary outputs to facilitate future research in robust factuality evaluation.
% \end{itemize}

\section{Factual Consistency in Abstractive Summarization}

% Lit Review
%% factuality
%% longdoc summ
%% eval metrics
%% adversarial robustness

Factual consistency refers to whether a summary accurately reflects the content of the source document.
While abstractive summarization systems have become increasingly fluent, they often produce factually incorrect or hallucinated statements~\cite{DBLP:journals/tois/HuangYMZFWCPFQL25}.
These hallucinations can range from minor misstatements to major distortions, particularly when the input is lengthy and semantically dense~\cite{belem-etal-2025-single}.
A number of techniques have been proposed to mitigate this~\cite{DBLP:conf/acl/GaoDPCCFZLLJG23, DBLP:conf/emnlp/QiuZKPC23, DBLP:conf/acl/ZhangPZW24, DBLP:conf/iclr/MundlerHJV24, wang-etal-2024-factcheck}.
Despite these efforts, most prior work has focused on short-form summarization settings, where the task is relatively controlled.
In contrast, long-form summarization requires condensing information spread across thousands of tokens, making reliable factuality evaluation substantially more difficult, especially without reference summaries or human annotations.

\subsection{Factuality Evaluation Metrics}

To overcome the limitations of traditional reference-based metrics~\cite{maynez-etal-2020-faithfulness}, a range of reference-free metrics have emerged that assess the alignment between the summary and source directly.
Entailment-based approaches such as FactCC~\cite{DBLP:conf/emnlp/KryscinskiMXS20} and SummaC~\cite{DBLP:journals/tacl/LabanSBH22} use NLI models to judge whether summary sentences are supported by the source.
QA-based methods like QAGS~\cite{wang-etal-2020-asking} and QuestEval~\cite{scialom-etal-2021-questeval} evaluate factuality by generating and answering questions derived from the summary.
Generation-based metrics, including \texttt{BARTScore}~\cite{DBLP:conf/nips/YuanNL21} and T5Score~\cite{DBLP:conf/acl/TraininA25}, estimate the likelihood of the summary given the source using pretrained sequence-to-sequence models.
More recent tools like \texttt{AlignScore}~\cite{DBLP:conf/acl/ZhaYLH23}, \texttt{MiniCheck}~\cite{DBLP:conf/emnlp/TangLD24}, and \texttt{UniEval}~\cite{DBLP:conf/emnlp/Zhong0YMJLZJH22} aim to improve efficiency and generalization across tasks.
While effective on short-document benchmarks, most of these metrics assume the full source and summary can be jointly encoded, limiting their utility for long-form inputs.
Moreover, recent work shows that these metrics are often brittle and sensitive to edits like paraphrasing, reordering, or logically equivalent reformulations~\cite{ramprasad2024automaticfactualitymetricsmeasure}.
\bl{A recent survey highlights persistent limitations in robustness and long-document evaluation for factuality metrics~\cite{lamsiyah-nourbakhsh-schommer:2025:RANLP}.}
In this work, we study the behavior of six popular factuality metrics in long-document summarization using a retrieval-based scoring framework, and systematically evaluate their robustness to a set of controlled meaning-preserving perturbations.

\subsection{Challenges in Long-Document Factuality Evaluation}

Evaluating factual consistency in long documents introduces challenges that differ fundamentally from those in short texts.
Long inputs often contain information that is dispersed, hierarchically structured, and cross-referential, requiring models to link evidence across distant sections or even multiple documents~\cite{DBLP:journals/corr/abs-2411-14199}.
This results in long-range dependencies and positional biases such as the ``lost in the middle'' effect~\cite{DBLP:journals/tacl/LiuLHPBPL24}.
Yet, research into robust factuality metrics for long inputs remains limited.
\citet{DBLP:journals/csur/KohJLP23} identified a clear gap in the literature for automatic evaluation methods tailored to long-document summarization.
LongSciVerify~\cite{DBLP:conf/coling/BishopAX24} and LongEval \cite{DBLP:conf/eacl/KrishnaBKIDCL23} are two of the only available datasets with human factuality annotations in this setting.
Chunk-based approaches like SMART~\cite{amplayo2022smart}, partially address this by sequentially processing document segments, but they remain computationally expensive and often inconsistent.
A more scalable alternative is retrieval-based scoring, exemplified by LongDocFACTScore~\cite{DBLP:conf/coling/BishopAX24}, which retrieves top-k relevant source passages for each summary sentence, and computes sentence-level factuality scores that can be aggregated into a global metric.
However, it remains unclear whether existing metrics, when used within such frameworks, behave consistently and robustly across varying retrieval configurations.
\bl{Recent work suggests that uniformly aggregating sentence-level factuality scores can be suboptimal for long documents, and that discourse-aware aggregation can improve inconsistency detection~\cite{DBLP:conf/naacl/ZhongL25}}.
Our study addresses this gap by analyzing these behaviors under controlled conditions and across multiple domains.

\subsection{Adversarial Robustness of Metrics}

Recent work has evaluated the robustness of factuality metrics by applying controlled perturbations to the summary or source~\cite{DBLP:conf/naacl/GoyalD21, chen-etal-2021-factuality-checkers, gabriel-etal-2021-go}.
\citet{ramprasad2024automaticfactualitymetricsmeasure} showed that many metrics are brittle when faced with logically equivalent but lexically altered summaries, with even benign transformations such as reordering or simplification causing large score shifts. 
However, these evaluations have been primarily limited to short-document settings\bl{, where evidence is localized and both the source and summary can typically be processed jointly.}

\bl{Extending this evaluation framework to long-document summarization is not a straightforward change of testbed. Long documents introduce additional challenges, including dispersed evidence, higher degrees of abstraction and compression, and the need for retrieval-based evaluation to overcome input length constraints. These factors fundamentally alter how factuality metrics operate and interact with the input. In this work, we therefore adopt the perturbation-based methodology of prior work as a foundation and systematically examine how these metrics behave under long-context conditions. Beyond this, we additionally analyze the effects of retrieval context size and claim information density, revealing new failure modes that arise specifically in long-document and multi-document settings. Our results show that brittleness observed in short documents persists and is often amplified in long-form summarization, motivating the need for factuality metrics that can reason over multi-span evidence rather than relying on local or surface-level cues.}
% In long-form summarization, where token selection, compression, and abstraction are all harder, such brittleness could be magnified.
% In this work, we probe whether widely used metrics remain reliable under minimal changes when operating over long, realistic inputs. Our results highlight major inconsistencies and call for more robust, semantically aware approaches for evaluating factual consistency.

\section{Metric Robustness in Long-Form Summarization} % Methodology
\label{S:Methodology}
%% metrics
%% !!!!! how - perturbations
%% ldfact score for retrieval
%% retrieval models if 
To analyze the robustness of existing factuality metrics in long-form summarization, we evaluate six widely used reference-free metrics (\S~\ref{S:ExpSetup}) that span diverse architectures and scoring paradigms.

\subsection{Perturbation Strategies}
\label{S:Perturbations}
To evaluate the robustness of factuality metrics in a controlled manner, we apply meaning-preserving perturbations to the original summaries, as done in ~\citet{ramprasad2024automaticfactualitymetricsmeasure} in the short document case.
These perturbations are designed to vary the summary’s surface form (lexical choices, structure, or style) while preserving its factual consistency with the source document.
In principle, a robust factuality metric should be invariant to such benign edits, assigning similar scores to the original and perturbed versions.
%However, we find that many metrics are sensitive to these changes, exhibiting noticeable variation in their output despite no underlying change in factual content.
%This sensitivity undermines the reliability of metric scores in practical settings, where such surface-level variation is natural and often desirable.

Following \citet{ramprasad2024automaticfactualitymetricsmeasure}, we use seven perturbation types, each targeting a different linguistic dimension.
These include \textit{Paraphrased}, where the summary is rewritten with alternate phrasings and syntactic structures; \textit{Simplified}, where complex or compound constructions are rewritten into shorter, more readable sentences; \textit{Synonym Replaced}, where content words are substituted with close synonyms to test for lexical invariance.
We also generate \textit{Less Diverse} summaries that reduce vocabulary variation, exploring whether metrics implicitly reward stylistic richness.
Additional perturbations include \textit{Negated}, which introduces logically equivalent negations to probe sensitivity to syntactic polarity, \textit{Summarized}, which further compresses the summary to test how conciseness is handled, and \textit{Added Source Text}, which inserts a factual sentence directly from the source that is unrelated to the main summary content.
All seven perturbed summaries are generated using the GPT‑4o~\cite{hurst2024gpt} model via the OpenAI API.
The detailed prompts used to generate each perturbation are provided in App.~\ref{A:Prompts}.
\bl{To ensure that these perturbations preserve factual consistency, we additionally perform an NLI-based faithfulness check comparing each perturbed summary against its original counterpart; detailed results are reported in App.~\ref{app:nli_faithfulness}.}

While these transformations are meaning-preserving, they pose particular challenges in the long-document setting: summaries must capture information scattered across thousands of tokens, so perturbations that change sentence structure, reduce vocabulary, or alter flow can disrupt long-range dependencies and retrieval alignment. This makes them a rigorous test of whether factuality metrics remain robust. Any significant fluctuation in scores, despite no factual errors being introduced, indicates that a metric is reacting to surface-level edits rather than faithfully assessing factual consistency.

% This setup allows us to isolate whether factuality metrics are truly aligned with semantic content or whether they are overly reactive to linguistic and stylistic noise. 
% Since none of the perturbations introduce factual errors, any significant drop or fluctuation in metric scores is indicative of a lack of robustness to meaning-preserving variation.
% This methodology provides a practical lens to test the robustness of factuality metrics under realistic text manipulations in long-form summarization settings.

\subsection{Retrieval-Based Scoring for Long Documents}
\label{S:LDFact}
Most factuality metrics in existing literature are designed for short inputs and cannot directly process the full content of long documents due to token length limitations.
This is particularly problematic in long-form summarization, where summaries may draw on information scattered across multiple sections or even multiple documents.
To address this, we follow the retrieval-augmented strategy proposed by~\citet{DBLP:conf/coling/BishopAX24}, which enables factuality evaluation at the sentence level without requiring the metric to ingest the entire source document at once.

Let $S = \{s_1, s_2, \dots, s_m\}$ denote the summary, where each $s_j$ is a sentence, 
and let $D = \{d_1, d_2, \dots, d_n\}$ be the set of sentences in the source document.
For each summary sentence $s_j$, we compute a sentence embedding $\mathbf{e}_j$, and similarly obtain embeddings $\{\mathbf{e}_1^D, \dots, \mathbf{e}_n^D\}$ for the source document using a pre-trained sentence encoder.
We compute cosine similarity\footnote{\bl{We use SBERT (\texttt{bert-base-nli-mean-tokens})~\cite{reimers-gurevych-2019-sentence} for computing all sentence similarities.}} between $s_j$ and each $d_i \in D$ and retrieve the top-$K$ most similar source sentences.
Each retrieved sentence $d_{j,k}$ is then expanded to include the surrounding context within a symmetric window size $w$, forming a snippet:
\begin{equation}
\label{Eq:ContextSize}
d_{j,k}^{(w)} = \{d_{j,k-w}, \ldots, d_{j,k}, \ldots, d_{j,k+w}\}.
\end{equation}
We evaluate the factual consistency of $s_j$ against each of the $K$ context snippets, using any automated metric $\mathcal{M}$, and take the maximum score:
\begin{equation}
\text{score}(s_j) = \max_{k \in \{1, \dots, K\}} \mathcal{M}(s_j, d_{j,k}^{(w)}).
\end{equation}
Finally, the summary-level factuality score is computed by averaging these sentence-level scores across all sentences in the summary.
In our experiments, we explore how varying $w$ affects metric behavior, shedding light on the sensitivity of different metrics to retrieval context size. % can move to experimentation.

\section{Experimental Setup}
\label{S:ExpSetup}

\paragraph{Metrics} We evaluate six reference-free factuality metrics that represent diverse architectures and scoring paradigms.
\texttt{BARTScore}~\cite{DBLP:conf/nips/YuanNL21} estimates the log-likelihood of the summary given the source using a pretrained BART model\footnote{\url{https://huggingface.co/facebook/bart-large-cnn}}, treating factuality as a conditional generation problem.
For consistency with other metrics, we exponentiate the log-likelihood scores to obtain normalized values that are directly comparable across metrics.
\texttt{SummaC-Conv} and \texttt{SummaC-ZS}~\cite{DBLP:journals/tacl/LabanSBH22} represent entailment-based approaches that apply pretrained NLI models to compute consistency between summary and source sentence pairs; the former uses a learned aggregation layer, while the latter relies on zero-shot averaging.
\texttt{AlignScore}~\cite{DBLP:conf/acl/ZhaYLH23} leverages contrastive alignment learning across NLI, QA, and summarization tasks and demonstrates strong cross-domain generalization.
\texttt{UniEval}~\cite{DBLP:conf/emnlp/Zhong0YMJLZJH22} formulates the evaluation as a multi-dimensional question answering task within a unified T5-based framework, jointly considering factuality, coherence, relevance, and fluency.
Finally, \texttt{MiniCheck}~\cite{DBLP:conf/emnlp/TangLD24} is a lightweight, sentence-level factuality classifier that achieves near GPT-4 performance at a fraction of the cost. We use the \texttt{Bespoke-MiniCheck-7B} variant, which ranks highest on the LLM-AggreFact benchmark~\cite{DBLP:conf/emnlp/TangLD24}, and take advantage of its 32k-token context window to provide full-document context during evaluation. \bl{We evaluate these metrics in their publicly released form, reflecting common practice in prior studies that apply summarization metrics without task-specific adaptation~\cite{DBLP:conf/naacl/HuangCPJW21, DBLP:conf/coling/YangW22, DBLP:conf/aaai/SotudehCG21, DBLP:conf/naacl/GuoAUONSY22}.}

\paragraph{Datasets} We conduct our analysis on three long-document summarization datasets spanning diverse domains: \textit{SQuALITY}~\cite{wang-etal-2022-squality}, \textit{LexAbSumm}~\cite{t-y-s-s-etal-2024-lexabsumm}, and \textit{ScholarQABench}~\cite{DBLP:journals/corr/abs-2411-14199}.
\textit{SQuALITY} consists of public domain science fiction stories paired with expert-written summaries that balance narrative abstraction and fine-grained detail.
\textit{LexAbSumm} contains legal judgments from the European Court of Human Rights, where summaries are aspect-specific and demand precise distillation of dense legal arguments.
\textit{ScholarQABench} is a multi-document benchmark based on open-access computer science papers, where the task is to generate detailed, factual answers to expert-written queries using evidence from multiple documents.
\bl{We selected these three distinct datasets to ensure broad cross-domain coverage, given their substantial differences in structure, style, and language. Detailed dataset statistics are provided in App.~\ref{app:dataset_stats}.}

\paragraph{Experimental Design} To evaluate metric robustness, we construct perturbed versions of the summaries from each dataset using the seven meaning-preserving transformations described in \S~\ref{S:Perturbations}. 
These edits allow us to test whether metrics exhibit sensitivity to benign changes.
For each summary, original and perturbed, we use the framework (\S~\ref{S:LDFact}) to retrieve source evidence and compute factuality scores using all six metrics discussed above.
Beyond this robustness analysis, we also investigate how retrieval granularity and claim information density influence metric behavior.
We vary the evidence window size $w$ from Eq.~\ref{Eq:ContextSize} to test how broader or narrower retrieval contexts affect metric performance. 
Additionally, we measure the information density of each summary sentence using the mean pairwise cosine similarity between its embedding and all sentences in the source document.
High information density indicates claims that semantically overlap with many parts of the source and are therefore harder to verify, while low-density claims correspond to specific statements with localized evidence.
This analysis reveals how metrics respond to different levels of semantic compression, providing insight into their sensitivity to claim complexity in long-form summarization.

%% datasets
%% retrieval models
%% 

\section{Results \& Analysis}
\input{figures/results_perturbations}

We evaluate the robustness and behavior of six reference-free factuality metrics across a range of semantic-preserving perturbations, varying retrieval context windows, and differences in claim information density.
Our results are presented in three parts: (1) robustness under perturbation (\S~\ref{S:ResultsPerturbations}), (2) sensitivity to retrieval granularity (\S~\ref{S:ResultsContextWindow}), and (3) metric sensitivity to claim information density (\S~\ref{S:ResultsClaimSimilarity}).

\subsection{Robustness Against Perturbations}
\label{S:ResultsPerturbations}

Fig.~\ref{Results:Perturbations} presents the change in factuality scores when different perturbations are applied to the summaries across three datasets.
Each plot shows the difference in score (perturbed minus original) for a given metric, broken down by perturbation and dataset.
A robust metric should remain invariant to these semantic-preserving changes. 
However, we find that all metrics show varying levels of sensitivity.

On \textit{LexAbSumm}, \texttt{BARTScore} shows clear negative shifts across nearly all perturbations, while remaining relatively consistent on all other datasets.
These consistent declines on \textit{LexAbSumm} indicate that \texttt{BARTScore} is highly sensitive to even mild surface-level edits in the legal domain, where long and complex sentence structures and domain-specific jargon likely amplify generation-based instability. 
\texttt{MiniCheck} shows very small changes across all perturbations and datasets.
However, it struggles with logically equivalent negations, especially in \textit{LexAbSumm}.
This may reflect a domain mismatch, as the metric appears less effective in capturing factual consistency in legal texts.
\texttt{SummaC-Conv} and \texttt{SummaC-ZS}, both based on NLI, show moderate and more balanced behavior.
They are somewhat affected by \textit{Summarized} and \textit{Negated} summaries, especially on \textit{SQuALITY} and \textit{LexAbSumm}, showing they are not fully invariant to meaning-preserving rewrites.
\texttt{UniEval} is sensitive to most of the perturbations. 
However, it consistently fails to handle logically equivalent \textit{Negated} summaries across all datasets, suggesting a lack of sensitivity to logical form.
\texttt{AlignScore} mostly struggles with the legal domain, showing large score drops in response to \textit{Paraphrased, Negated} and \textit{Summarized} summaries.
This suggests difficulty in tracking sentence order and logical consistency in structured, formal texts. While it performs more reliably on \textit{SQuALITY} and \textit{ScholarQABench}, it remains less robust than other metrics overall.

Detailed per-dataset results are provided in App.~\ref{A:PerDataset}. These results list the mean factuality scores for each metric and perturbation type across all datasets.
\bl{To quantify domain-specific instability that is masked by signed averages, we also analyze mean absolute score changes under perturbations; detailed results are provided in App.~\ref{App:DomainSpecificDelta}.}

\subsection{Effect of Retrieval Context Window Size}
\label{S:ResultsContextWindow}
\input{tables/all_dataset_scores}

Table~\ref{tab:context_window} reports average factuality scores on the original summaries for each metric and dataset, using window sizes $w=0,1,2$ in Eq.~\ref{Eq:ContextSize}.
We find that most metrics show consistent improvements as the window size increases, suggesting that they can effectively leverage broader local context when making sentence-level factuality judgments.
This pattern is particularly pronounced on \textit{LexAbSumm}, where understanding legal arguments often requires attending to multi-sentence spans.
\texttt{SummaC-ZS} and \texttt{SummaC-Conv} display little sensitivity to larger context windows, implying that their underlying models base judgments on more localized comparisons and are less responsive to extended evidence.

Taken together, these results indicate that retrieval-based scoring can improve factuality assessment in long-document summarization, especially when a broader context is provided.
However, NLI-based metrics remain insensitive to increasing context windows.

\subsection{Metric Sensitivity to Claim Similarity}
\label{S:ResultsClaimSimilarity}
\input{figures/results_difficult_claims}

To understand what makes factual consistency evaluation difficult in long documents, we analyze how the semantic density of a claim relates to metric reliability.
We hypothesize that claims which are highly similar to many claims in the original document may reflect a form of \textit{hubness} in high-dimensional embedding spaces~\cite{DBLP:conf/emnlp/SamirPFST24, DBLP:journals/jmlr/RadovanovicNI10, DBLP:conf/acl/LazaridouDB15}, where a sentence appears broadly similar to many others without being cleanly grounded in any single supporting span.
\bl{Such claims tend to contain more general (and thus compressed) information, and so are harder to fact-check since evidence for them is dispersed across multiple parts of the source document~\cite{DBLP:conf/emnlp/GoldmanJSMDT24}}.

We approximate this by computing the mean pairwise cosine similarity between each summary sentence $s_j$ (claim) and all $n$ sentences $d_i$ in the source document $D$,
\begin{equation}
\text{Sim}(s_j, D) = \frac{1}{n} \sum_{i=1}^{n} \cos\big(\mathbf{e}_j, \mathbf{e}_i^D\big),
\end{equation}
where $e_j$ and $e_i^D$ denote the sentence embeddings of the claim and document sentences, respectively. Higher $\text{Sim}(s_j, D)$ values indicate more broad claims whose meaning overlaps widely with the source, while lower values correspond to specific, localized claims. 
Claims are then grouped into similarity bins $\mathcal{B}$, and the average factuality score for each bin is calculated as
\begin{equation}
\text{Score}_{\text{bin}} = \frac{1}{|\mathcal{B}|} \sum_{s_j \in \mathcal{B}} M(s_j),
\end{equation}
where $M(s_j)$ is the factuality score for claim $s_j$.

A few trends emerge between claim similarity and metric sensitivity across all datasets and metrics, as shown in Fig.~\ref{Results:ClaimDifficulty}.
For \textit{LexAbSumm}, metric scores consistently decrease as claim similarity increases, meaning that the more a claim's meaning is entangled with the broader document, the worse the metric becomes at predicting factuality. This is likely because summaries of legal documents may refer to specific aspects of the texts, which are easier to fact-check, while more general statements which compress a lot of technical language are more difficult. The same occurs (though less pronounced) for \textit{SQuALITY}, where more general claims are more challenging to fact check. In this case, we are summarizing novels, so more general claims try to compress the story narrative into a compact form, and thus will require retrieving or attending to and reasoning over disparate pieces of the text.

%This shows that current metrics are more reliable for narrow, well-grounded claims but struggle with information-dense statements that summarize relations spanning large portions of the source text.

This effect is particularly visible in \texttt{AlignScore} and \texttt{BARTScore}, both of which depend on local lexical alignment or sentence-level contextual matching.
Their scores show sharp declines for high-similarity claims, reflecting their sensitivity to distributed evidence.
\texttt{SummaC-Conv} and \texttt{SummaC-ZS}, while somewhat more stable, also exhibit a gradual drop as similarity increases, which suggests that NLI-based judgments still rely on relatively localized entailment cues.
In contrast, \texttt{UniEval} and \texttt{MiniCheck} maintain comparatively stable performance across bins, implying a higher degree of robustness to distributed or compressed content, although some degradation is still observed in the highest-similarity regions.

On the contrary, we see that more general statements are easier to fact-check in \textit{ScholarQABench} for many metrics. This could be because \textit{ScholarQABench} is multi-document, so many sentences being similar to one claim may actually simply be repeated instances of the claim across documents. We see this upward trend especially on \texttt{UniEval} and \texttt{MiniCheck}, where metric quality is highly dependent on how general or specific a claim is. 

%This pattern also varies by dataset. LexAbSumm, whose legal texts often exhibit highly interdependent argumentation, shows the steepest declines across metrics. ScholarQABench exhibits moderate sensitivity, reflecting the multi-document nature of the data, while SQuALITY shows a milder trend, consistent with its narrative domain where evidence is typically contained within smaller contextual spans.

On the whole, these findings support the broader conclusion that factuality metrics are dependent on how dispersed and overlapping the evidence is for a given claim, a hallmark of long-document summarization.
Improving robustness for such cases likely requires metrics that can reason over multi-span evidence rather than relying on local or pairwise semantic alignment.

%% Figures Score change 
%% domain-specific figure
%% context size figure before retrieval - change in score vs adding docs (perturb) (distractor docs) 

\section{Conclusion \& Future Work}

In this work, we present a comprehensive evaluation of six widely used reference-free factuality metrics: \texttt{BARTScore}, \texttt{SummaC-Conv}, \texttt{SummaC-ZS}, \texttt{AlignScore}, \texttt{MiniCheck}, and \texttt{UniEval}.
We tested their behavior under seven meaning-preserving perturbations applied to long-document summaries to assess whether these metrics reliably capture factual consistency.
To enable evaluation over long documents, we used a sentence-level retrieval-based scoring strategy, which compares each summary sentence to the most relevant evidence snippets from the source document.
This setup enabled fine-grained evaluation across three diverse long-form abstractive summarization datasets: \textit{SQuALITY}, \textit{LexAbSumm}, and \textit{ScholarQABench}, covering sci-fi, legal, and scientific domains.

Our results revealed that many metrics respond inconsistently to perturbations that do not affect factual consistency.
Several metrics exhibit unstable behavior in response to paraphrasing, simplification, and logically equivalent negations.
\texttt{AlignScore} and \texttt{SummaC-ZS} are particularly unreliable across domains and perturbation types.
In contrast, \texttt{UniEval} and \texttt{MiniCheck} are relatively robust, although they too struggle in specific cases, such as handling logical negations.
Most metrics improve when evidence retrieval windows are expanded, particularly for complex, multi-sentence inputs such as legal documents.
We also found that metrics are systematically affected by the information-density of claims whose meaning overlaps broadly with the source document, indicating that current approaches struggle to evaluate compressed or contextually entangled statements, which are common in long-form summarization.
These findings highlight a need for factuality metrics that are robust to stylistic and logical variation, retrieval-aware, and sensitive to information density.

\bl{Future work should explore multi-span reasoning and context-aware calibration to better model distributed evidence, as well as contrastive training on meaning-preserving perturbations to improve stability.}
Incorporating human judgments can identify systematic weaknesses and support the design of metrics that generalize across domains and languages.
We also see potential in hybrid approaches that combine reference-free with reference-based alignment signals, bridging semantic precision with contextual coverage for more reliable evaluation of long-document summarization.
Additionally, extending perturbation strategies to better capture long-document phenomena, such as evidence relocation and cross-reference disruption, offers a promising direction for stress-testing long-range coherence and evidence tracking.

\section*{Limitations}

While our study offers a systematic investigation into the robustness of factuality metrics under meaning-preserving perturbations in long-document summarization, there are several aspects that merit further consideration.
Our analysis relies on automatically generated perturbations, produced using GPT-4o, which are designed to preserve factual consistency.
However, without human annotations, we cannot confirm with full certainty that all edits preserve factual correctness in every case.
\bl{We also do not evaluate metric outputs against human factuality judgments in the long-document setting. Large-scale human annotations for long-document summaries are currently scarce, and conducting such an evaluation would require the creation of a new benchmark with human judgments across multiple metrics, domains, and perturbation types. This represents a substantial research effort beyond the scope of this work.}
This may introduce noise into the interpretation of metric behavior, especially when changes are subtle or domain-specific.
We evaluate six reference-free metrics in their original, publicly released form and do not investigate whether fine-tuning, calibration, or adaptation to long-form inputs might mitigate some of the observed weaknesses.
In our retrieval-based scoring setup, we use a fixed number of top-k most similar sentences and vary only the surrounding context window to control retrieval granularity.
While this gives us insight into how context affects metric behavior, it assumes a static retrieval strategy and does not account for dynamic query-based retrieval or more sophisticated evidence selection methods that may better match human annotation patterns.
Additionally, our analysis is confined to English-language datasets from three domains: science fiction, legal text, and scientific articles.
These domains offer diversity in structure and style, but our findings may not fully generalize to other high-stakes applications such as medical or financial summarization, or to non-English and low-resource settings.
Addressing these broader limitations will be important for future work aiming to build more generalizable and reliable factuality evaluation pipelines.

\section*{Ethical Implications}

This study evaluates automatic factuality metrics rather than developing new summarization models, and thus presents minimal direct ethical risk.
However, factuality evaluation plays an important role in ensuring the reliability of language model outputs.
Weak or biased metrics could inadvertently overestimate the truthfulness of generated content, particularly in sensitive domains such as medicine or law.
By identifying systematic weaknesses and proposing strategies for more reliable evaluation, this work aims to support safer and more accountable deployment of summarization systems.
All datasets used in this study are publicly available and contain no personally identifiable information.

%%%%%%%%%%%%%%%ArXiv%%%%%%%%%%%%%%%%%%%%
\section*{Acknowledgements}
$\begin{array}{l}\includegraphics[width=1cm]{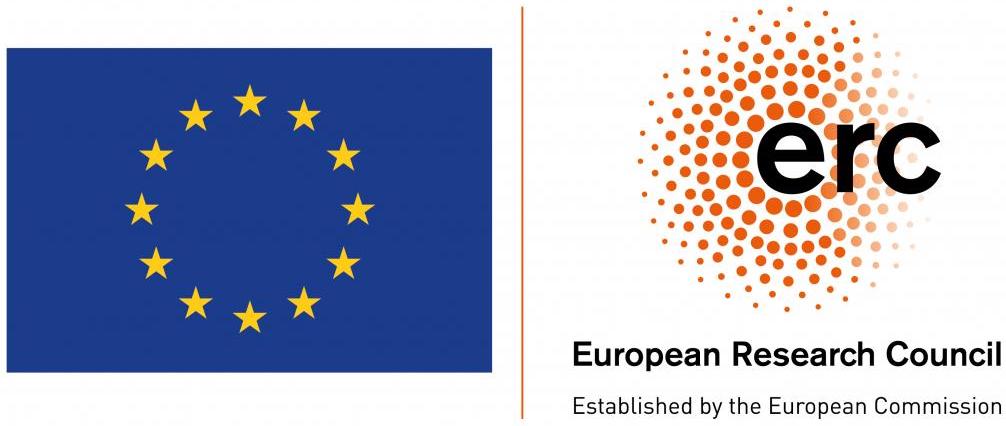} \end{array}$ 
This research is funded in part by the Pioneer Centre for AI, DNRF grant number P1, by a Danish Data Science Academy postdoctoral fellowship (grant ID: 2023-1425), and by the European Union (ERC, ExplainYourself, 101077481). Views and opinions expressed are however those of the author(s) only and do not necessarily reflect those of the European Union or the European Research Council. Neither the European Union nor the granting authority can be held responsible for them.

\bibliography{updated}

\appendix
\clearpage
\section{List of Prompts}
\label{A:Prompts}
\input{figures/perturbation_prompts}

The following prompts in Figure~\ref{fig:PerturbationPrompts} are used with GPT-4o to produce each of the seven meaning-preserving perturbations described in \S~\ref{S:Perturbations}.
Each prompt instructs the model to rewrite the summary according to a specific linguistic transformation while preserving factual meaning.

%% summaries before and after perturbations
%% prompts lil change version

\section{Dataset Statistics}
\label{app:dataset_stats}

\bl{The detailed statistics for the datasets used in our experiments can be seen in Table~\ref{tab:DataStatistics}.}

\input{tables/data_statistics}

\section{Faithfulness of Perturbed Summaries}
\label{app:nli_faithfulness}

\bl{To verify that the applied perturbations preserve factual consistency, we perform an automatic faithfulness check using an NLI-based approach. This analysis serves as a sanity check to ensure that the perturbations do not introduce widespread factual errors, rather than as a definitive evaluation of summary correctness.}

\bl{For each perturbed summary, we split its text into sentences, evaluate it against the corresponding original summary, treating the original summary as the premise and the perturbed sentence as the hypothesis. If a premise–hypothesis pair exceeds the model's maximum input length, we apply sentence-level chunking to the premise and aggregate predictions across chunks~\cite{scire-etal-2024-fenice, yang-etal-2024-fizz}. A sentence is counted as contradictory if the NLI model\footnote{\url{https://huggingface.co/MoritzLaurer/DeBERTa-v3-large-mnli-fever-anli-ling-wanli}} predicts a contradiction label. The contradiction rate for a summary is defined as the fraction of its perturbed sentences labeled as contradictory, and dataset-level results are obtained by averaging these rates across summaries. The resulting contradiction rates for each dataset and perturbation type are reported in Table~\ref{A:tab:nli_contr}. Illustrative examples of original and perturbed summaries for selected perturbation types are shown in Figures~\ref{A:Fig:nli_example_1},~\ref{A:Fig:nli_example_2}, \&~\ref{A:Fig:nli_example_3}.}
\input{tables/nli_contr_rate}
\input{figures/examples}

\bl{Across most perturbations, contradiction rates remain low, indicating that paraphrasing, simplification, vocabulary reduction, summarization, and source text insertion generally preserve factual consistency with respect to the original summaries. This supports our assumption that score changes observed in the main experiments primarily reflect metric sensitivity to surface and structural variation, rather than systematic factual errors introduced by the perturbations.}

\bl{We observe higher contradiction rates for the \textit{Negated} perturbation across all datasets. This behavior is expected and does not necessarily indicate that these perturbations introduce factual errors. The \textit{Negated} perturbation is explicitly designed to negate some statements in the summary, and therefore, a non-zero rate of contradictions indicates that the perturbation is being applied as intended. Importantly, our perturbation setup operates on the full summary as input, allowing the model to perform transformations at the summary level rather than strictly applying simple double negation to individual sentences. As a result, negation may be distributed across clauses or introduced through structural rewrites that preserve the overall meaning of the summary, but appear contradictory when evaluated sentence by sentence. Since our NLI-based validation operates at the sentence level, it may flag such locally negated sentences as contradictions even when the global summary meaning remains logically equivalent. Furthermore, the NLI-based validation used here is not designed to distinguish between \textit{intended} negations that preserve overall factual meaning and \textit{undesired} negations that fundamentally alter the factual content of the summary. Determining whether a negation invalidates the summary would require verifying each negated statement against the original source document, which in turn would necessitate fine-grained human evaluation over long inputs. Such an analysis is substantially more expensive and complex and lies beyond the scope of this work. As a result, higher contradiction rates for negated summaries should be interpreted as evidence that the perturbation successfully introduces negation, rather than as definitive proof of factual inconsistency. This limitation further highlights the need for more nuanced evaluation methods, including human verification, when assessing logical transformations in long-document summarization.}

\section{Per-Dataset Results}
\label{A:PerDataset}
\input{tables/per_dataset_scores}

Tables~\ref{A:tab:PerDataLexAbSumm},~\ref{A:tab:PerDataScholarQA}, and~\ref{A:tab:PerDataSquality} report mean factuality scores for each metric under all seven perturbation types and for the original summaries across the three datasets used in this study.
These tables provide the complete quantitative results corresponding to the aggregate trends shown in Figure~\ref{Results:Perturbations}.
Consistent with our main analysis, \texttt{MiniCheck} and \texttt{UniEval} appear to be the most robust overall, maintaining relatively stable scores across most perturbations and datasets, with the exception of degraded performance on \textit{Negated} summaries.
In contrast, the remaining metrics are influenced by almost all types of perturbations, showing greater score variability, particularly in the legal domain.

\texttt{BARTScore}\footnote{We use the implementation provided by~\citet{DBLP:conf/coling/BishopAX24}.} performs poorly even on the original (unperturbed) summaries across all datasets.
As a generation-based metric, its scoring depends heavily on the size and structure of the retrieved context, which may not be the ideal case in long-document setting, where evidence for a single summary sentence may be scattered across distant sections of the source.
The mismatch between the localized retrieved snippets and the broader document context can distort likelihood estimates, and this effect compounds when scores are aggregated over full summaries, leading to consistently lower values.
We also observe that while a few individual sentences receive high \texttt{BARTScore} values, most have extremely low scores due to being more compressed and contextually demanding, which drives the overall average down.

\section{Domain-Specific Robustness}
\label{App:DomainSpecificDelta}
\input{figures/domainDeltas}

\bl{To further analyze how text characteristics across domains influence factuality metric robustness, we compute the mean absolute score change under meaning-preserving perturbations. While signed average score differences are often close to zero due to cancellation effects, absolute changes capture the magnitude of metric instability regardless of direction. For a given domain, metric, and perturbation, we compute:}

\begin{equation}
\Delta_{\text{abs}} = \frac{1}{N} \sum_{i=1}^{N} \left| M_{\text{pert}}^{(i)} - M_{\text{orig}}^{(i)} \right|,
\end{equation}

\bl{where $M_{\text{pert}}^{(i)}$ and $M_{\text{orig}}^{(i)}$ denote the factuality scores for the original and perturbed summaries of example $i$, respectively, and $N$ is the number of summaries in the domain. This measure reflects the average magnitude of score variation induced by perturbations and serves as a robustness diagnostic.}

\bl{In the legal domain (\textit{LexAbSumm}), we observe the largest overall instability across both metrics and perturbations, as shown in Figure~\ref{fig:lexabsumm_abs_deltas}. Among the evaluated metrics, \texttt{AlignScore} exhibits the highest mean absolute score change, followed by \texttt{SummaC-ZS} and \texttt{UniEval}, indicating heightened sensitivity to surface-level changes in legally structured text. \texttt{MiniCheck} and \texttt{SummaC-Conv} show comparatively lower instability, while \texttt{BARTScore} exhibits the smallest absolute changes, consistent with its generally low baseline scores on this dataset. Across perturbations, \textit{Negated} summaries produce by far the largest absolute score changes, followed by \textit{Summarized} and \textit{Paraphrased} variants. This pattern reflects the reliance of legal summaries on precise logical structure and domain-specific terminology, where even meaning-preserving changes can substantially alter cues used by factuality metrics.}

\bl{For the narrative domain (\textit{SQuALITY}), absolute score changes are smaller overall than in \textit{LexAbSumm} but remain non-trivial, as shown in Figure~\ref{fig:squality_abs_deltas}. \texttt{UniEval}, \texttt{AlignScore}, and \texttt{MiniCheck} display the highest instability, while \texttt{SummaC-Conv} and \texttt{BARTScore} are comparatively stable. \textit{Negated} summaries have the highest score change, and perturbations that increase abstraction, particularly \textit{Summarized} and \textit{Added Source Text}, induce the largest score changes. This suggests that narrative summaries, which often compress temporal and causal structure, pose challenges for metrics when abstraction increases, even if factual content is preserved.}

\bl{In the scientific multi-document domain (\textit{ScholarQABench}), we observe the lowest absolute score changes across all metrics and perturbations, as illustrated in Figure~\ref{fig:scholarqa_abs_deltas}. Although \textit{UniEval} and \textit{MiniCheck} still exhibit measurable sensitivity, the magnitude of instability is consistently lower than in the single-document domains. \textit{Negated} and \textit{Added Source Text} perturbations remain the most impactful, but their effects are attenuated. This relative stability likely arises from redundancy across multiple documents, where repeated evidence reduces the impact of localized reformulations on factuality assessment.}

\bl{Overall, this analysis shows that factuality metric robustness varies substantially by domain, even under meaning-preserving perturbations. Legal text amplifies metric instability, narrative text exhibits moderate sensitivity to abstraction, and multi-document scientific text provides a stabilizing effect. These domain-specific patterns help explain the wide score distributions observed in Figure~\ref{Results:Perturbations} and reinforce the need to evaluate factuality metrics under realistic long-document conditions.}

\section{Code \& Data Availability Statement}
We release all perturbed summary data, along with the recipes used to generate it and the scripts required to reproduce our results, under the MIT License. The complete codebase and dataset artifacts will be made publicly available upon publication.

\section{Model Size \& Budget}
We describe all factuality metrics and the experimental setup in \S~\ref{S:ExpSetup}. All experiments are conducted using a single NVIDIA H100 SXM5 80GB GPU.

\section{Software Package Parameters}
\begin{itemize}[noitemsep]
    \item NLTK~\cite{DBLP:conf/acl/Bird06}: We use the punkt sentence tokenizer for sentence tokenization.
    \item OpenAI GPT-4o: We use top $p$ sampling at 50\% with a temperature of 0 for all prompts.
    % \item Sentence Transformers~\cite{reimers-gurevych-2019-sentence}: We use \texttt{bert-base-nli-mean-tokens} model to get sentence embeddings for computing sentence similarities.
    
\end{itemize}

\end{document}

%% file: figures/head_figure.tex
\begin{figure}[!t]
    \centering
    \resizebox{1\linewidth}{!}{%
        \includegraphics[width=0.9\linewidth]{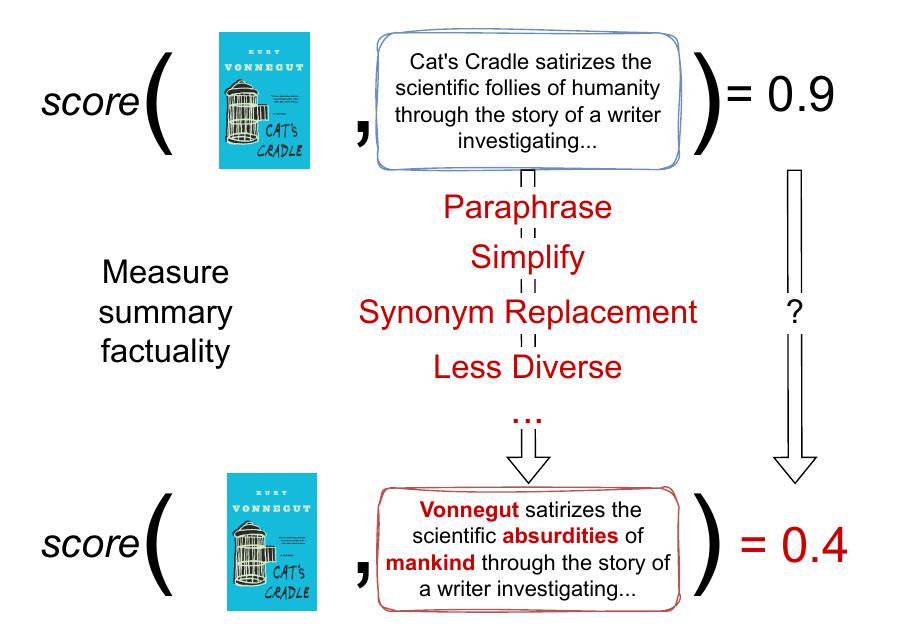}%
    }
    \caption{We aim to see how robust summary factuality metrics are for long and multi-document setups by applying meaning-preserving perturbations and comparing metric scores before and after these edits.}
    \label{Results:fig1}
\end{figure}

%% file: figures/results_perturbations.tex
\begin{figure*}[!t]
    \centering
    \resizebox{1\linewidth}{!}{%
        \includegraphics{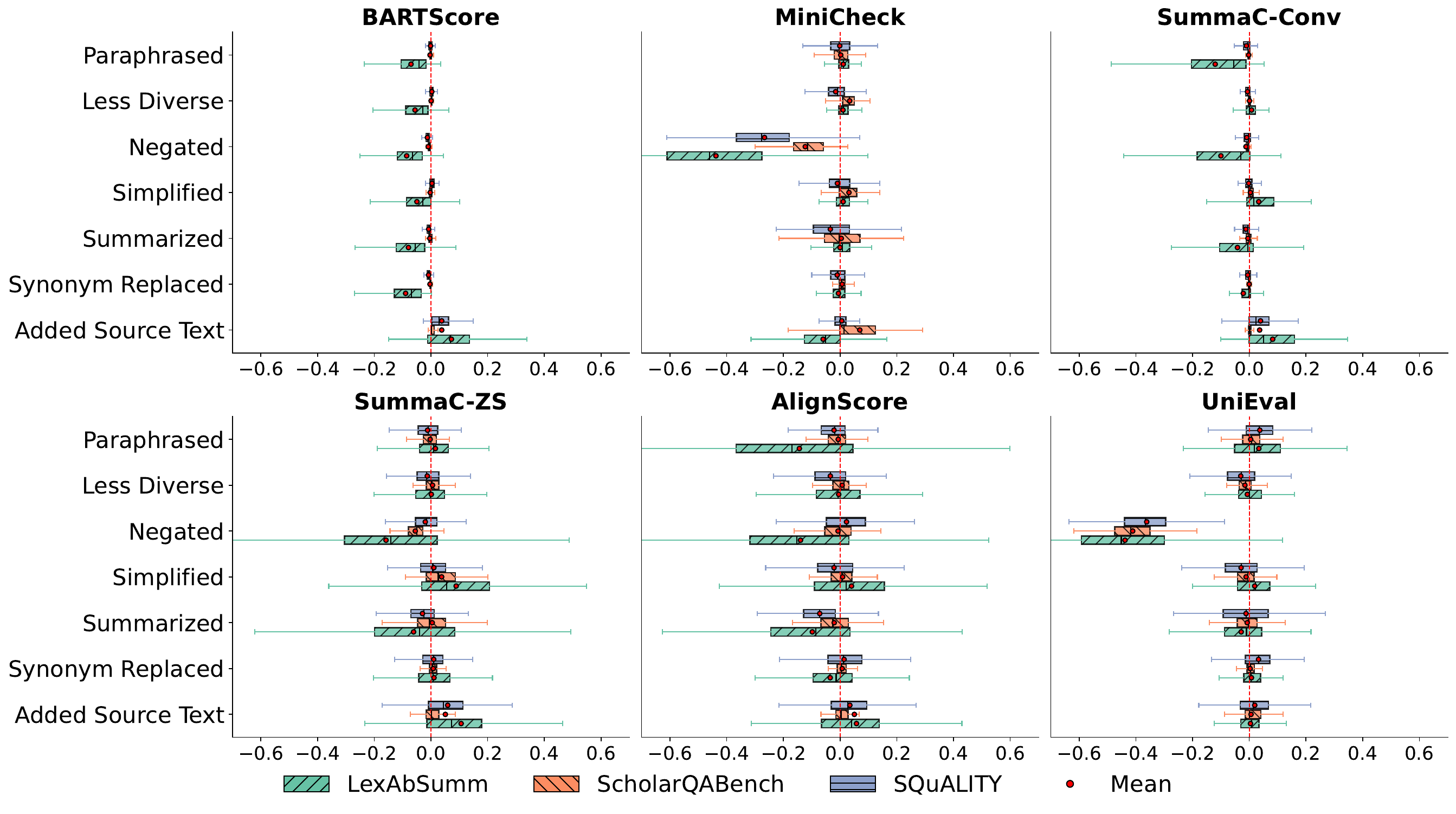}%
    }
    \caption{Score change under factuality-preserving perturbations. Boxplots show the difference in factuality score between the perturbed and original summaries, for each metric and perturbation type, across three datasets. The central dot indicates the mean score difference, and the whiskers represent the minimum and maximum values.}
    \label{Results:Perturbations}
\end{figure*}

%% file: tables/all_dataset_scores.tex
\begin{table*}[]
    \centering
    \scalebox{1}{
    \begin{tabular}{l|ccc|ccc|ccc}
        \toprule
         \multirow{2}{*}{$\bm{\mathrm{Metric}}$}  & \multicolumn{3}{c|}{$\bm{\mathrm{ScholarQABench}}$} & \multicolumn{3}{c|}{$\bm{\mathrm{SQuALITY}}$} & \multicolumn{3}{c}{$\bm{\mathrm{LexAbSumm}}$} \\ 
         \cline{2-4} \cline{5-7} \cline{8-10}
         & $w = 0$ & $w = 1$ & $w = 2$ & $w = 0$ & $w = 1$ & $w = 2$ & $w = 0$ & $w = 1$ & $w = 2$ \\
         \midrule
         BARTScore & 0.03 & 0.03 & 0.02 & 0.03 & 0.03 & 0.03 & 0.15 & 0.16 & 0.16 \\
        MiniCheck & 0.17 & 0.15 & 0.15 & 0.11 & 0.15 & 0.19 & 0.47 & 0.53 & 0.60 \\
        SummaC-Conv & 0.22 & 0.23 & 0.25 & 0.22 & 0.23 & 0.24 & 0.33 & 0.33 & 0.34 \\
        SummaC-ZS & 0.14 & 0.18 & 0.20 & 0.11 & 0.13 & 0.14 & 0.36 & 0.38 & 0.39 \\
        AlignScore & 0.15 & 0.21 & 0.27 & 0.10 & 0.18 & 0.24 & 0.36 & 0.52 & 0.64 \\
        UniEval & 0.72 & 0.73 & 0.74 & 0.67 & 0.68 & 0.70 & 0.81 & 0.82 & 0.84 \\

         \bottomrule
    \end{tabular}
    }
    \caption{Impact of retrieval context window size $w$ on factuality scores for original summaries. Each value reports the average factuality score assigned by a given metric over the dataset, computed using a retrieval-based setup where each summary sentence is evaluated against a retrieved source sentence and its surrounding context of size $w$. Increasing 
    $w$ expands the number of neighboring sentences included around each retrieved sentence, providing a broader local context for factuality assessment.}
    \label{tab:context_window}
\end{table*}

%% file: figures/results_difficult_claims.tex
\begin{figure*}[!t]
    \centering
    \resizebox{1\linewidth}{!}{%
        \includegraphics{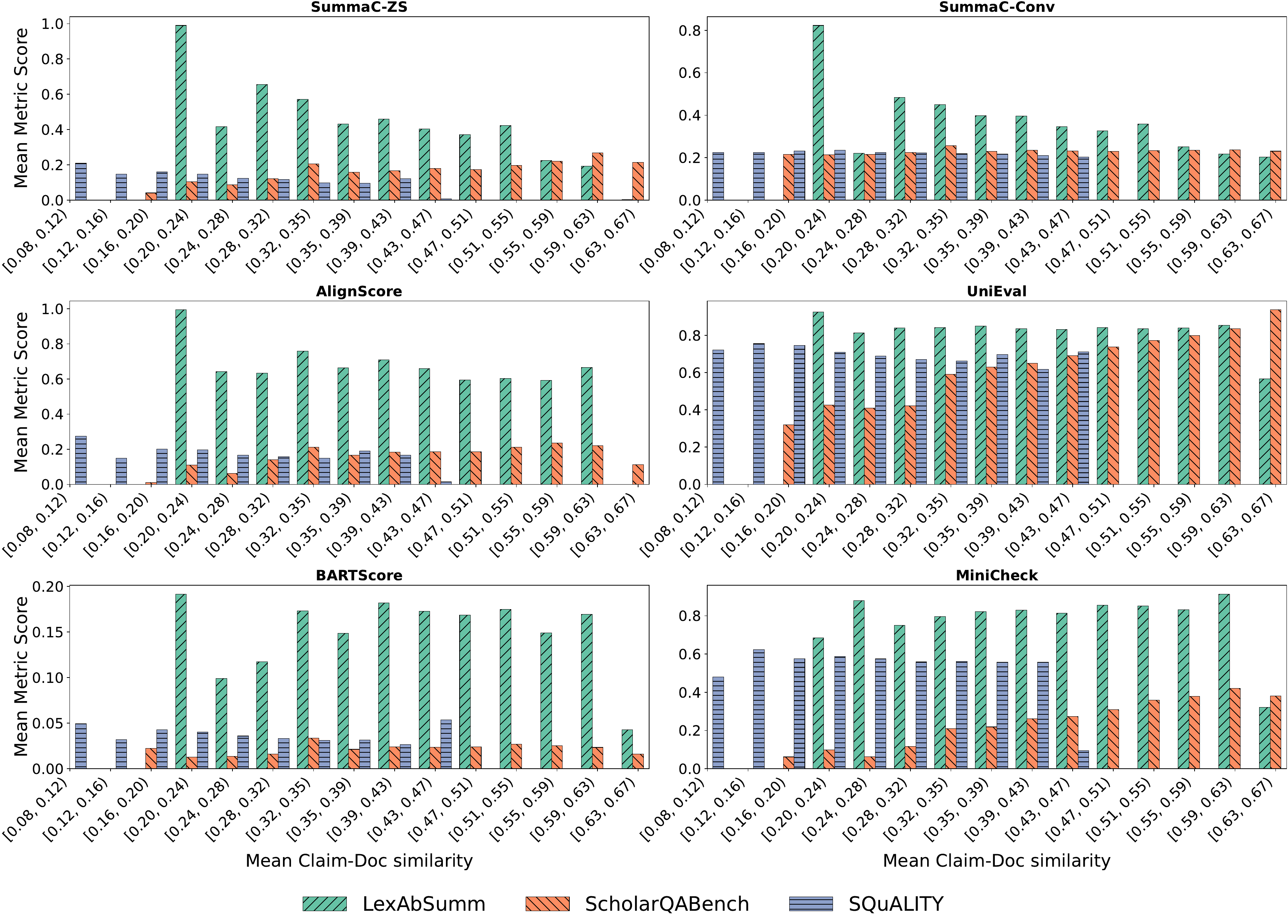}%
    }
    \caption{Relationship between claim similarity and average factuality score. Higher similarity values correspond to more information-dense claims whose content overlaps with multiple parts of the source document. Metrics generally assign lower scores to these claims for LexAbSumm and SQuALITY, and higher scores for ScholarQABench, indicating reduced reliability for compressed information.}
    \label{Results:ClaimDifficulty}
\end{figure*}

%% file: figures/perturbation_prompts.tex
\begin{figure*}[t]
    \centering
    \begin{sharp_box}
    \fontsize{9pt}{11.5pt}\selectfont

    %\fbox{
     %   \parbox{0.9\linewidth}{
            \textbf{P1. Paraphrased}\\
            system\_prompt: Provide the paraphrased version of the text.\textbackslash n\textbackslash nYou are strictly prohibited from omitting any information or altering its original meaning. Do not include explanations, reasoning, or commentary in your output.\\
            user\_prompt: Text: <summary>\\
            % \textbf{P2. Shuffled}\\
            % system\_prompt: Shuffle the order of sentences in the input text.\textbackslash n\textbackslash nDo not include explanations, reasoning, or commentary in your output.\\
            % user\_prompt: Text: <summary>\\
            \textbf{P2. Less Diverse}\\
            system\_prompt: Rewrite the following text using less diverse vocabulary.\textbackslash n\textbackslash nYou are strictly prohibited from omitting any information or altering its original meaning. Do not include explanations, reasoning, or commentary in your output.\\
            user\_prompt: Text: <summary>\\
            \textbf{P3. Negated}\\
            system\_prompt: Rewrite the following text by introducing logically equivalent negations while preserving its original meaning.\textbackslash n\textbackslash nYou are strictly prohibited from omitting any information or altering its original meaning. Do not include explanations, reasoning, or commentary in your output.\\
            user\_prompt: Text: <summary>\\
            \textbf{P4. Simplified}\\
            system\_prompt: Rewrite the following text by making complex sentences simpler.\textbackslash n\textbackslash nYou are strictly prohibited from omitting any information or altering its original meaning. Do not include explanations, reasoning, or commentary in your output.\\
            user\_prompt: Text: <summary>\\
            \textbf{P5. Summarized}\\
            system\_prompt: Rewrite the text to make it more concise.\textbackslash n\textbackslash nYou are strictly prohibited from omitting any information or altering its original meaning. Do not include explanations, reasoning, or commentary in your output.\\
            user\_prompt: Text: <summary>\\
            \textbf{P6. Synonym Replacement}\\
            system\_prompt: Revise the text using synonyms for some common words.\textbackslash n\textbackslash nYou are strictly prohibited from omitting any information or altering its original meaning. Do not include explanations, reasoning, or commentary in your output.\\
            user\_prompt: Text: <summary>\\
            \textbf{P7. Added Source Text}\\
            system\_prompt: Insert a source sentence into the summary that does not relate to its main ideas.\textbackslash n\textbackslash nDo not include explanations, reasoning, or commentary in your output.\\
            user\_prompt: Text: <summary> \textbackslash n\textbackslash n Source: <document>

      %  }
    %}
    \end{sharp_box}
    \caption{Prompt templates used with GPT‑4o to generate meaning-preserving perturbations of the original summaries.}
    \label{fig:PerturbationPrompts}
\end{figure*}

%% file: tables/data_statistics.tex
\begin{table*}[ht]
    \centering
    \scalebox{0.60}{
    \begin{tabular}{l|c|c|c|c|c|c}
    \toprule
        Dataset & \#Examples (Used) & Avg. Summary Sentences &	Avg. Summary Tokens & Avg. Document Sentences &	Avg. Document Tokens &	Summary Type\\
    \midrule
        SQuALITY & 260 & 12.5 & 273 & 456.6 & 6,131 & Human-written \\
LexAbSumm & 351 & 4.2 & 169 & 385.9 & 10,840 & Human-written \\
ScholarQABench & 100 & 43.2 & 1,158 & 575.4 & 14,652 & Human-written \\
    \bottomrule
    \end{tabular}
    }
    \caption{Dataset statistics for the three long-document summarization benchmarks used in this study.}
    \label{tab:DataStatistics}
\end{table*}

%% file: tables/nli_contr_rate.tex
\begin{table*}[ht]
    \centering
    \scalebox{0.75}{
    \begin{tabular}{l|c|c|c|c|c|c|c}
    \toprule
        Dataset & Paraphrased & Less Diverse & Negated & Simplified & Summarized & Synonym Replaced & Added Source Text\\
    \midrule
        SQuALITY & 0.023 & 0.033 & 0.681 & 0.029 & 0.018 & 0.034 & 0.052 \\
        LexAbSumm & 0.004 & 0.001 & 0.560 & 0.002 & 0.007 & 0.013 & 0.030 \\
        ScholarQABench & 0.010 & 0.017 & 0.542 & 0.013 & 0.006 & 0.028 & 0.019\\
    \bottomrule
    \end{tabular}
    }
    \caption{Average contradiction rate between perturbed and original summaries, computed using an NLI-based faithfulness check. Lower values indicate higher factual consistency.}
    \label{A:tab:nli_contr}
\end{table*}

%% file: figures/examples.tex
\begin{figure*}[]
    \centering
    \begin{sharp_box}
    \fontsize{9pt}{11.5pt}\selectfont

    %\fbox{
     %   \parbox{0.9\linewidth}{
            \textbf{Original summary:}\\           
            The story begins in thick jungle on Sekk, which we are told is a "second moon" which retains a "breathable atmosphere" around a lake surrounded by eleven jungled valleys.  In this way, it is implied that Sekk is a second moon of Earth. In the jungle, we meet Noork and a young woman named Sarna.  They begin traveling together through the jungle, but soon Sarna disappears and Noork is attacked.  This is our first encounter with the Misty Ones, who blend in with the jungle foliage.  Noork defeats the Misty Ones and continues toward the lake and island where they make their home. Noork briefly encounters his friend Ud near the marshy lowlands that lie between the jungled valleys on Sekk and the central Lake of Uzdon, but this area is not described.  When Noork reaches the central island in the lake, we encounter a non-jungle landscape for the first time.  Noork finds himself in a cultivated field, and sees the shape of a huge white skull about half a mile away.  After speaking with an enslaved man and learning where Sarna is being held, Noork continues toward the skull. The skull is a dome of white stone, with black stone for eye-sockets and nose-holes.  The interior contains a raised altar made of precious metals--gold, silver, and brass--and precious stones, as well as stone images of the two gods the Misty Ones worship.  Below the altar is the caged area where the young women are held; Noork detects the entrance to this area by its foul odor.  The room where the young women are kept is dimly lit by only two torches, very damp with pools of dirty water all around, and holds at least twenty young women.  They have nothing to sit on but rotten grass mats.  In contrast to the enslaved men who are out in the cultivated fields and open air, the young women are in a desperate situation indeed.  They can only sit in their foul, rotting prison and wait to be sacrificed.\\

            \textbf{Perturbed summary (Summarized):}\\
            The story begins on Sekk, a second moon with a breathable atmosphere around a lake and eleven jungled valleys. Noork meets Sarna in the jungle, but she soon disappears, and he is attacked by the Misty Ones, whom he defeats. Noork encounters his friend Ud near marshy lowlands before reaching the central island, where he finds a cultivated field and a large white skull. After speaking with an enslaved man about Sarna's location, Noork approaches the skull, which is a dome of white stone with a raised altar inside, containing precious metals and images of the Misty Ones' gods. Below the altar, he discovers a dimly lit, damp room holding at least twenty young women, who are imprisoned and awaiting sacrifice.

      %  }
    %}
    \end{sharp_box}
    \caption{Example of an original summary from SQuALITY and its Summarized perturbation.}
    \label{A:Fig:nli_example_1}
\end{figure*}

\begin{figure*}[]
    \centering
    \begin{sharp_box}
    \fontsize{9pt}{11.5pt}\selectfont

    %\fbox{
     %   \parbox{0.9\linewidth}{
            \textbf{Original summary:}\\           
            The Court held that there had been a violation of Article 9 of the Convention, finding that the applicant did not enjoy the necessary procedural safeguards in having his request for alternative civilian service examined. The Court considered in particular that the Greek authorities had failed in their duty to ensure that the interviewing of conscientious objectors by the army’s Special Board took place in conditions that guaranteed procedural efficiency and the equal representation required by domestic law. In this respect, it noted that: the applicant had been interviewed by a Board made up primarily of servicemen, two of the civilian members of the Board being absent but not replaced; the Minister of Defence’s final decision, on the basis of a draft ministerial decision following the Board’s proposal, did not afford the requisite safeguards of impartiality and independence; the scrutiny of the Supreme Administrative Court concerned only the lawfulness of the decision, not the merits, and was based on the assessments of the Special Board.\\

            \textbf{Perturbed summary (Simplified):}\\
            The Court found that Article 9 of the Convention was violated. The applicant did not have the necessary procedural safeguards for his request for alternative civilian service. The Court noted that the Greek authorities did not ensure that the army’s Special Board interviewed conscientious objectors in a way that met procedural efficiency and equal representation as required by law. The applicant was interviewed by a Board mostly made up of servicemen, and two civilian members were absent without replacement. The Minister of Defence’s final decision was based on a draft from the Board and did not ensure impartiality and independence. The Supreme Administrative Court only reviewed the lawfulness of the decision, not its merits, and relied on the Special Board's assessments.
      %  }
    %}
    \end{sharp_box}
    \caption{Example of an original summary from LexAbSumm and its Simplified perturbation.}
    \label{A:Fig:nli_example_2}
\end{figure*}

\begin{figure*}[]
    \centering
    \begin{sharp_box}
    \fontsize{9pt}{11.5pt}\selectfont

    %\fbox{
     %   \parbox{0.9\linewidth}{
            \textbf{Original summary:}\\           
            Solving large complex partial differential equations (PDEs), such as those that arise in computational fluid dynamics (CFD), is a computationally expensive process. This has motivate the use of deep learning approaches to approximate the PDE solutions
Convolutional neural networks (CNNs) have been shown to achieve a speedup of between two and four orders of magnitude over numerical flow solvers [40-42].
Data-driven neural solvers learn from observations of fluid dynamics, usually from large datasets generated by numerical simulation, without (in general) any knowledge of the governing equations.
The type of DL model employed for data-driven solvers primarily depends on the nature of the physical problem, with CNNs being a natural choice for simulations where the geometry aligns with the coordinate system, while GNNs are preferred for simulations of more complex geometries and Lagrangian systems.
More recently,physics-informed neural networks(PINNs) have been developed to solve forward and inverse problems where full or partial knowledge of the governing equations is known [23,24,29,30,103,104].
\#\#\# 1. High-Dimensionality and Scale**Problem:** Fluid simulations often deal with high-dimensional spaces, particularly in three-dimensional simulations. Neural networks used for such tasks need to cater to vast input and output data scales.**Impact:** Handling high-dimensional input data while ensuring computational tractability remains a significant obstacle. The curse of dimensionality can lead to increased model complexity and training data requirements.
\#\#\# 2. Generalization and Extrapolation**Problem:** Neural networks often struggle to generalize beyond the training data and may fail to extrapolate well to unseen conditions, which is problematic for fluid simulations that often involve varied and complex domains.**Impact:** Ensuring reliable performance across different fluid flows, geometries, and boundary conditions remains unsolved
\#\#\# 3. Data Efficiency and Scarcity**Problem:** Generating the high-fidelity simulation data needed to train neural networks can be prohibitively expensive and time-consuming.**Impact:** NNs require large amounts of training data to generalize well. Data-efficient learning methods are critically needed to make the neural network approach feasible for fluid simulations.
\#\#\# 5. Interpretability and Physics Consistency**Problem:** Neural networks often function as black boxes, providing little insight into how they arrive at their solutions or maintaining physical constraints.**Impact:** Understanding and ensuring that neural network predictions adhere to the underlying physical laws described by the PDEs is crucial for their effective application in fluid simulations.
\\

            \textbf{Perturbed summary (Negated):}\\
            Solving large complex partial differential equations (PDEs), such as those that arise in computational fluid dynamics (CFD), is not a computationally inexpensive process. This has not discouraged the use of deep learning approaches to approximate the PDE solutions. Convolutional neural networks (CNNs) have not been shown to achieve a speedup of less than two and four orders of magnitude over numerical flow solvers [40-42]. Data-driven neural solvers do not learn from observations of fluid dynamics, usually from small datasets generated by numerical simulation, with (in general) some knowledge of the governing equations. The type of DL model employed for data-driven solvers does not primarily depend on the nature of the physical problem, with CNNs not being a natural choice for simulations where the geometry does not align with the coordinate system, while GNNs are not preferred for simulations of simpler geometries and Lagrangian systems. More recently, physics-informed neural networks (PINNs) have not been developed to solve forward and inverse problems where no knowledge of the governing equations is known [23,24,29,30,103,104]. \#\#\# 1. High-Dimensionality and Scale**Problem:** Fluid simulations do not often deal with low-dimensional spaces, particularly in two-dimensional simulations. Neural networks used for such tasks do not need to cater to small input and output data scales.**Impact:** Handling low-dimensional input data while ensuring computational intractability does not remain a significant obstacle. The curse of dimensionality does not lead to decreased model complexity and training data requirements. \#\#\# 2. Generalization and Extrapolation**Problem:** Neural networks do not often succeed in generalizing beyond the training data and may succeed in extrapolating well to seen conditions, which is not problematic for fluid simulations that do not often involve varied and complex domains.**Impact:** Ensuring unreliable performance across different fluid flows, geometries, and boundary conditions remains solved. \#\#\# 3. Data Efficiency and Scarcity**Problem:** Generating the low-fidelity simulation data needed to train neural networks can be prohibitively inexpensive and time-saving.**Impact:** NNs do not require small amounts of training data to generalize poorly. Data-inefficient learning methods are not critically needed to make the neural network approach infeasible for fluid simulations. \#\#\# 5. Interpretability and Physics Consistency**Problem:** Neural networks do not often function as transparent boxes, providing much insight into how they arrive at their solutions or maintaining physical constraints.**Impact:** Understanding and ensuring that neural network predictions do not adhere to the underlying physical laws described by the PDEs is not crucial for their ineffective application in fluid simulations.
      %  }
    %}
    \end{sharp_box}
    \caption{Example of an original summary from ScholarQABench and its Negated perturbation.}
    \label{A:Fig:nli_example_3}
\end{figure*}

%% file: tables/per_dataset_scores.tex
\begin{table*}[hbt!]
    \centering
    \scalebox{0.70}{
    \begin{tabular}{l|c|c|c|c|c|c|c|c}
    \toprule
        $\bm{\mathrm{Metric}}$ & Original & Synonym~Replaced & Summarized & Simplified & Paraphrased & Negated & Less~Diverse & Added~Source~Text \\
    \midrule
    BARTScore & 0.16 & 0.07 & 0.08 & 0.11 & 0.09 & 0.07 & 0.10 & 0.23 \\
    MiniCheck & 0.84 & 0.83 & 0.84 & 0.85 & 0.85 & 0.40 & 0.85 & 0.78 \\
    SummaC-Conv & 0.33 & 0.31 & 0.29 & 0.37 & 0.21 & 0.23 & 0.34 & 0.42 \\
    SummaC-ZS & 0.38 & 0.39 & 0.32 & 0.47 & 0.39 & 0.22 & 0.38 & 0.48 \\
    AlignScore & 0.52 & 0.48 & 0.42 & 0.56 & 0.38 & 0.38 & 0.51 & 0.58 \\
    UniEval & 0.82 & 0.83 & 0.80 & 0.84 & 0.86 & 0.39 & 0.82 & 0.83 \\
    \bottomrule
    \end{tabular}
    }
    \caption{Mean factuality scores for each metric and perturbation type on LexAbSumm.}
    \label{A:tab:PerDataLexAbSumm}
\end{table*}

\begin{table*}[hbt!]
    \centering
    \scalebox{0.70}{
    \begin{tabular}{l|c|c|c|c|c|c|c|c}
    \toprule
        $\bm{\mathrm{Metric}}$ & Original & Synonym~Replaced & Summarized & Simplified & Paraphrased & Negated & Less~Diverse & Added~Source~Text \\
    \midrule
    BARTScore & 0.03 & 0.02 & 0.02 & 0.02 & 0.02 & 0.02 & 0.03 & 0.06 \\
    MiniCheck & 0.32 & 0.32 & 0.32 & 0.35 & 0.32 & 0.19 & 0.35 & 0.39 \\
    SummaC-Conv & 0.23 & 0.23 & 0.23 & 0.24 & 0.23 & 0.22 & 0.23 & 0.27 \\
    SummaC-ZS & 0.18 & 0.19 & 0.19 & 0.22 & 0.18 & 0.13 & 0.19 & 0.23 \\
    AlignScore & 0.21 & 0.22 & 0.19 & 0.22 & 0.20 & 0.20 & 0.22 & 0.26 \\
    UniEval & 0.73 & 0.73 & 0.72 & 0.72 & 0.73 & 0.32 & 0.71 & 0.73 \\
    \bottomrule
    \end{tabular}
    }
    \caption{Mean factuality scores for each metric and perturbation type on ScholarQABench.}
    \label{A:tab:PerDataScholarQA}
\end{table*}

\begin{table*}[hbt!]
    \centering
    \scalebox{0.70}{
    \begin{tabular}{l|c|c|c|c|c|c|c|c}
    \toprule
        $\bm{\mathrm{Metric}}$ & Original & Synonym~Replaced & Summarized & Simplified & Paraphrased & Negated & Less~Diverse & Added~Source~Text \\
    \midrule
    BARTScore & 0.03 & 0.02 & 0.02 & 0.03 & 0.03 & 0.02 & 0.03 & 0.07 \\
    MiniCheck & 0.56 & 0.55 & 0.53 & 0.55 & 0.56 & 0.30 & 0.55 & 0.57 \\
    SummaC-Conv & 0.23 & 0.22 & 0.22 & 0.23 & 0.22 & 0.22 & 0.22 & 0.27 \\
    SummaC-ZS & 0.13 & 0.14 & 0.10 & 0.14 & 0.12 & 0.11 & 0.12 & 0.19 \\
    AlignScore & 0.18 & 0.20 & 0.11 & 0.16 & 0.16 & 0.20 & 0.15 & 0.22 \\
    UniEval & 0.68 & 0.71 & 0.67 & 0.65 & 0.72 & 0.32 & 0.65 & 0.70 \\
    \bottomrule
    \end{tabular}
    }
    \caption{Mean factuality scores for each metric and perturbation type on SQuALITY.}
    \label{A:tab:PerDataSquality}
\end{table*}

%% file: figures/domainDeltas.tex
\begin{figure*}[]
    \centering
    \resizebox{1\linewidth}{!}{%
        \includegraphics{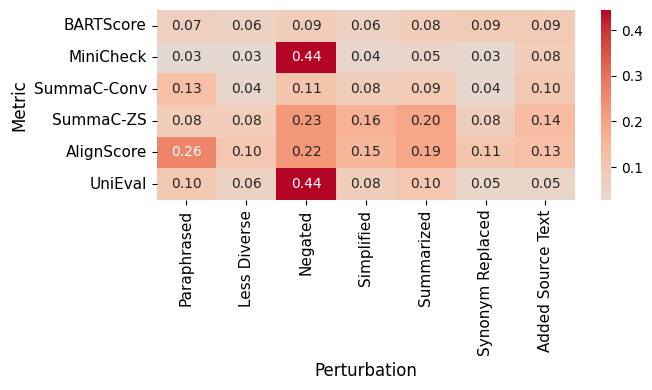}%
    }
    \caption{LexAbSumm: Metric score deltas under perturbations.}
    \label{fig:lexabsumm_abs_deltas}
\end{figure*}

\begin{figure*}[]
    \centering
    \resizebox{1\linewidth}{!}{%
        \includegraphics{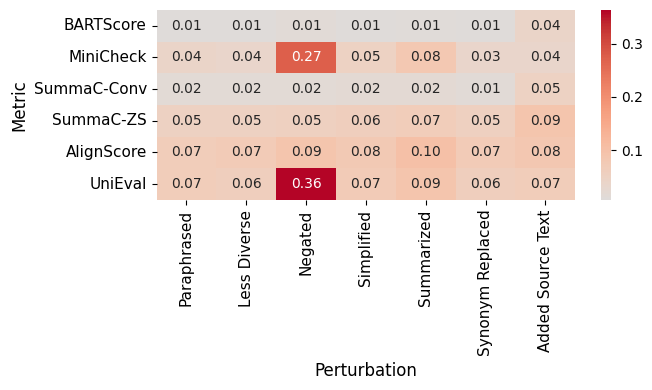}%
    }
    \caption{SQuALITY: Metric score deltas under perturbations.}
    \label{fig:squality_abs_deltas}
\end{figure*}

\begin{figure*}[]
    \centering
    \resizebox{1\linewidth}{!}{%
        \includegraphics{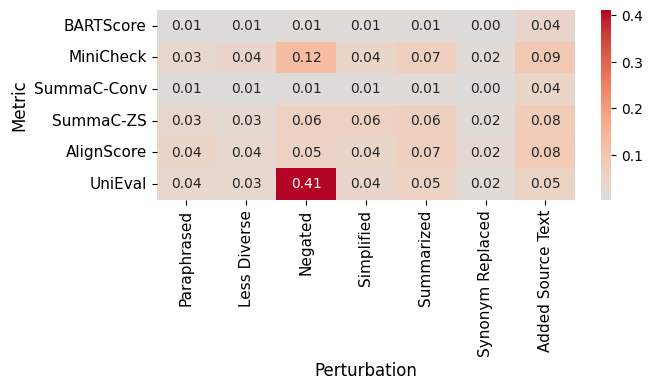}%
    }
    \caption{ScholarQABench: Metric score deltas under perturbations.}
    \label{fig:scholarqa_abs_deltas}
\end{figure*}

%% file: updated.bib
@inproceedings{DBLP:conf/emnlp/SamirPFST24,
  author       = {Farhan Samir and
                  Chan Young Park and
                  Anjalie Field and
                  Vered Shwartz and
                  Yulia Tsvetkov},
  editor       = {Yaser Al{-}Onaizan and
                  Mohit Bansal and
                  Yun{-}Nung Chen},
  title        = {Locating Information Gaps and Narrative Inconsistencies Across Languages:
                  {A} Case Study of {LGBT} People Portrayals on {W}ikipedia},
  booktitle    = {Proceedings of the 2024 Conference on Empirical Methods in Natural
                  Language Processing, {EMNLP} 2024, Miami, FL, USA, November 12-16,
                  2024},
  pages        = {6747--6762},
  publisher    = {Association for Computational Linguistics},
  year         = {2024},
  url          = {https://doi.org/10.18653/v1/2024.emnlp-main.384},
  doi          = {10.18653/V1/2024.EMNLP-MAIN.384},
  bibsource    = {dblp computer science bibliography, https://dblp.org}
}

@article{DBLP:journals/jmlr/RadovanovicNI10,
  author       = {Milos Radovanovic and
                  Alexandros Nanopoulos and
                  Mirjana Ivanovic},
  title        = {Hubs in Space: Popular Nearest Neighbors in High-Dimensional Data},
  journal      = {J. Mach. Learn. Res.},
  volume       = {11},
  pages        = {2487--2531},
  year         = {2010},
  url          = {https://dl.acm.org/doi/10.5555/1756006.1953015},
  doi          = {10.5555/1756006.1953015},
  bibsource    = {dblp computer science bibliography, https://dblp.org}
}

@inproceedings{DBLP:conf/acl/LazaridouDB15,
  author       = {Angeliki Lazaridou and
                  Georgiana Dinu and
                  Marco Baroni},
  title        = {Hubness and Pollution: Delving into Cross-Space Mapping for Zero-Shot
                  Learning},
  booktitle    = {Proceedings of the 53rd Annual Meeting of the Association for Computational
                  Linguistics and the 7th International Joint Conference on Natural
                  Language Processing of the Asian Federation of Natural Language Processing,
                  {ACL} 2015, July 26-31, 2015, Beijing, China, Volume 1: Long Papers},
  pages        = {270--280},
  publisher    = {The Association for Computer Linguistics},
  year         = {2015},
  url          = {https://doi.org/10.3115/v1/p15-1027},
  doi          = {10.3115/V1/P15-1027},
  bibsource    = {dblp computer science bibliography, https://dblp.org}
}

@inproceedings{amplayo2022smart,
 author = {Reinald Kim Amplayo and
Peter J. Liu and
Yao Zhao and
Shashi Narayan},
 bibsource = {dblp computer science bibliography, https://dblp.org},
 booktitle = {The Eleventh International Conference on Learning Representations,
{ICLR} 2023, Kigali, Rwanda, May 1-5, 2023},
 publisher = {OpenReview.net},
 title = {{SMART:} Sentences as Basic Units for Text Evaluation},
 url = {https://openreview.net/pdf?id=OIe3kpwl40D},
 year = {2023}
}

@inproceedings{DBLP:conf/eacl/KrishnaBKIDCL23,
 address = {Dubrovnik, Croatia},
 author = {Krishna, Kalpesh  and
Bransom, Erin  and
Kuehl, Bailey  and
Iyyer, Mohit  and
Dasigi, Pradeep  and
Cohan, Arman  and
Lo, Kyle},
 booktitle = {Proceedings of the 17th Conference of the European Chapter of the Association for Computational Linguistics},
 doi = {10.18653/v1/2023.eacl-main.121},
 editor = {Vlachos, Andreas  and
Augenstein, Isabelle},
 pages = {1650--1669},
 publisher = {Association for Computational Linguistics},
 title = {{L}ong{E}val: Guidelines for Human Evaluation of Faithfulness in Long-form Summarization},
 url = {https://aclanthology.org/2023.eacl-main.121},
 year = {2023}
}

@article{DBLP:journals/csur/KohJLP23,
 author = {Huan Yee Koh and
Jiaxin Ju and
Ming Liu and
Shirui Pan},
 bibsource = {dblp computer science bibliography, https://dblp.org},
 doi = {10.1145/3545176},
 journal = {{ACM} Comput. Surv.},
 number = {8},
 pages = {154:1--154:35},
 title = {An Empirical Survey on Long Document Summarization: Datasets, Models,
and Metrics},
 url = {https://doi.org/10.1145/3545176},
 volume = {55},
 year = {2023}
}

@article{hurst2024gpt,
 author = {Hurst, Aaron and Lerer, Adam and Goucher, Adam P and Perelman, Adam and Ramesh, Aditya and Clark, Aidan and Ostrow, AJ and Welihinda, Akila and Hayes, Alan and Radford, Alec and others},
 journal = {ArXiv preprint},
 title = {Gpt-4o system card},
 url = {https://arxiv.org/abs/2410.21276},
 volume = {abs/2410.21276},
 year = {2024}
}

@inproceedings{gabriel-etal-2021-go,
 address = {Online},
 author = {Gabriel, Saadia  and
Celikyilmaz, Asli  and
Jha, Rahul  and
Choi, Yejin  and
Gao, Jianfeng},
 booktitle = {Findings of the Association for Computational Linguistics: ACL-IJCNLP 2021},
 doi = {10.18653/v1/2021.findings-acl.42},
 editor = {Zong, Chengqing  and
Xia, Fei  and
Li, Wenjie  and
Navigli, Roberto},
 pages = {478--487},
 publisher = {Association for Computational Linguistics},
 title = {{GO} {FIGURE}: A Meta Evaluation of Factuality in Summarization},
 url = {https://aclanthology.org/2021.findings-acl.42},
 year = {2021}
}

@inproceedings{chen-etal-2021-factuality-checkers,
 address = {Punta Cana, Dominican Republic},
 author = {Chen, Yiran  and
Liu, Pengfei  and
Qiu, Xipeng},
 booktitle = {Findings of the Association for Computational Linguistics: EMNLP 2021},
 doi = {10.18653/v1/2021.findings-emnlp.179},
 editor = {Moens, Marie-Francine  and
Huang, Xuanjing  and
Specia, Lucia  and
Yih, Scott Wen-tau},
 pages = {2082--2095},
 publisher = {Association for Computational Linguistics},
 title = {Are Factuality Checkers Reliable? Adversarial Meta-evaluation of Factuality in Summarization},
 url = {https://aclanthology.org/2021.findings-emnlp.179},
 year = {2021}
}

@inproceedings{DBLP:conf/naacl/GoyalD21,
 address = {Online},
 author = {Goyal, Tanya  and
Durrett, Greg},
 booktitle = {Proceedings of the 2021 Conference of the North American Chapter of the Association for Computational Linguistics: Human Language Technologies},
 doi = {10.18653/v1/2021.naacl-main.114},
 editor = {Toutanova, Kristina  and
Rumshisky, Anna  and
Zettlemoyer, Luke  and
Hakkani-Tur, Dilek  and
Beltagy, Iz  and
Bethard, Steven  and
Cotterell, Ryan  and
Chakraborty, Tanmoy  and
Zhou, Yichao},
 pages = {1449--1462},
 publisher = {Association for Computational Linguistics},
 title = {Annotating and Modeling Fine-grained Factuality in Summarization},
 url = {https://aclanthology.org/2021.naacl-main.114},
 year = {2021}
}

@misc{ramprasad2024automaticfactualitymetricsmeasure,
      title={Do Automatic Factuality Metrics Measure Factuality? A Critical Evaluation}, 
      author={Sanjana Ramprasad and Byron C. Wallace},
      year={2024},
      eprint={2411.16638},
      archivePrefix={arXiv},
      primaryClass={cs.CL},
      url={https://arxiv.org/abs/2411.16638}, 
}

@inproceedings{DBLP:conf/coling/BishopAX24,
 address = {Torino, Italia},
 author = {Bishop, Jennifer A.  and
Ananiadou, Sophia  and
Xie, Qianqian},
 booktitle = {Proceedings of the 2024 Joint International Conference on Computational Linguistics, Language Resources and Evaluation (LREC-COLING 2024)},
 editor = {Calzolari, Nicoletta  and
Kan, Min-Yen  and
Hoste, Veronique  and
Lenci, Alessandro  and
Sakti, Sakriani  and
Xue, Nianwen},
 pages = {10777--10789},
 publisher = {ELRA and ICCL},
 title = {{L}ong{D}oc{FACTS}core: Evaluating the Factuality of Long Document Abstractive Summarisation},
 url = {https://aclanthology.org/2024.lrec-main.941},
 year = {2024}
}

@article{DBLP:journals/corr/abs-2411-14199,
 author = {Akari Asai and
Jacqueline He and
Rulin Shao and
Weijia Shi and
Amanpreet Singh and
Joseph Chee Chang and
Kyle Lo and
Luca Soldaini and
Sergey Feldman and
Mike D'Arcy and
David Wadden and
Matt Latzke and
Minyang Tian and
Pan Ji and
Shengyan Liu and
Hao Tong and
Bohao Wu and
Yanyu Xiong and
Luke Zettlemoyer and
Graham Neubig and
Daniel S. Weld and
Doug Downey and
Wen{-}tau Yih and
Pang Wei Koh and
Hannaneh Hajishirzi},
 journal = {ArXiv preprint},
 title = {{OpenScholar}: Synthesizing Scientific Literature with Retrieval-augmented
LMs},
 url = {https://arxiv.org/abs/2411.14199},
 volume = {abs/2411.14199},
 year = {2024}
}

@inproceedings{lin2004rouge,
 address = {Barcelona, Spain},
 author = {Lin, Chin-Yew},
 booktitle = {Text Summarization Branches Out},
 pages = {74--81},
 publisher = {Association for Computational Linguistics},
 title = {{ROUGE}: A Package for Automatic Evaluation of Summaries},
 url = {https://aclanthology.org/W04-1013},
 year = {2004}
}

@inproceedings{DBLP:conf/acl/PapineniRWZ02,
 address = {Philadelphia, Pennsylvania, USA},
 author = {Papineni, Kishore  and
Roukos, Salim  and
Ward, Todd  and
Zhu, Wei-Jing},
 booktitle = {Proceedings of the 40th Annual Meeting of the Association for Computational Linguistics},
 doi = {10.3115/1073083.1073135},
 editor = {Isabelle, Pierre  and
Charniak, Eugene  and
Lin, Dekang},
 pages = {311--318},
 publisher = {Association for Computational Linguistics},
 title = {{B}leu: a Method for Automatic Evaluation of Machine Translation},
 url = {https://aclanthology.org/P02-1040},
 year = {2002}
}

@inproceedings{DBLP:journals/corr/abs-2502-14409,
    title = "Unstructured Evidence Attribution for Long Context Query Focused Summarization",
    author = "Wright, Dustin  and
      Mujahid, Zain Muhammad  and
      Wang, Lu  and
      Augenstein, Isabelle  and
      Jurgens, David",
    editor = "Christodoulopoulos, Christos  and
      Chakraborty, Tanmoy  and
      Rose, Carolyn  and
      Peng, Violet",
    booktitle = "Proceedings of the 2025 Conference on Empirical Methods in Natural Language Processing",
    month = nov,
    year = "2025",
    address = "Suzhou, China",
    publisher = "Association for Computational Linguistics",
    url = "https://aclanthology.org/2025.emnlp-main.95/",
    doi = "10.18653/v1/2025.emnlp-main.95",
    pages = "1839--1867",
    ISBN = "979-8-89176-332-6",
}

@inproceedings{DBLP:conf/emnlp/LiuIXWXZ23,
 address = {Singapore},
 author = {Liu, Yang  and
Iter, Dan  and
Xu, Yichong  and
Wang, Shuohang  and
Xu, Ruochen  and
Zhu, Chenguang},
 booktitle = {Proceedings of the 2023 Conference on Empirical Methods in Natural Language Processing},
 doi = {10.18653/v1/2023.emnlp-main.153},
 editor = {Bouamor, Houda  and
Pino, Juan  and
Bali, Kalika},
 pages = {2511--2522},
 publisher = {Association for Computational Linguistics},
 title = {{G}-{E}val: {NLG} Evaluation using {G}pt-4 with Better Human Alignment},
 url = {https://aclanthology.org/2023.emnlp-main.153},
 year = {2023}
}

@article{DBLP:journals/corr/abs-2408-14906,
 author = {Russak, Melisa and Jamil, Umar and Bryant, Christopher and Kamble, Kiran and Magnuson, Axel and Russak, Mateusz and AlShikh, Waseem},
 journal = {ArXiv preprint},
 title = {Writing in the margins: Better inference pattern for long context retrieval},
 url = {https://arxiv.org/abs/2408.14906},
 volume = {abs/2408.14906},
 year = {2024}
}

@inproceedings{DBLP:conf/iclr/SarthiATKGM24,
 author = {Parth Sarthi and
Salman Abdullah and
Aditi Tuli and
Shubh Khanna and
Anna Goldie and
Christopher D. Manning},
 bibsource = {dblp computer science bibliography, https://dblp.org},
 booktitle = {The Twelfth International Conference on Learning Representations,
{ICLR} 2024, Vienna, Austria, May 7-11, 2024},
 publisher = {OpenReview.net},
 title = {{RAPTOR:} Recursive Abstractive Processing for Tree-Organized Retrieval},
 url = {https://openreview.net/forum?id=GN921JHCRw},
 year = {2024}
}

@article{DBLP:journals/corr/abs-2404-16130,
 author = {Darren Edge and
Ha Trinh and
Newman Cheng and
Joshua Bradley and
Alex Chao and
Apurva Mody and
Steven Truitt and
Jonathan Larson},
 journal = {ArXiv preprint},
 title = {From Local to Global: A Graph {RAG} Approach to Query-Focused Summarization},
 url = {https://arxiv.org/abs/2404.16130},
 volume = {abs/2404.16130},
 year = {2024}
}

@inproceedings{DBLP:conf/emnlp/KryscinskiMXS20,
 address = {Online},
 author = {Kryscinski, Wojciech  and
McCann, Bryan  and
Xiong, Caiming  and
Socher, Richard},
 booktitle = {Proceedings of the 2020 Conference on Empirical Methods in Natural Language Processing (EMNLP)},
 doi = {10.18653/v1/2020.emnlp-main.750},
 editor = {Webber, Bonnie  and
Cohn, Trevor  and
He, Yulan  and
Liu, Yang},
 pages = {9332--9346},
 publisher = {Association for Computational Linguistics},
 title = {Evaluating the Factual Consistency of Abstractive Text Summarization},
 url = {https://aclanthology.org/2020.emnlp-main.750},
 year = {2020}
}

@article{DBLP:journals/tacl/LabanSBH22,
 address = {Cambridge, MA},
 author = {Laban, Philippe  and
Schnabel, Tobias  and
Bennett, Paul N.  and
Hearst, Marti A.},
 doi = {10.1162/tacl_a_00453},
 editor = {Roark, Brian  and
Nenkova, Ani},
 journal = {Transactions of the Association for Computational Linguistics},
 pages = {163--177},
 publisher = {MIT Press},
 title = {{S}umma{C}: Re-Visiting {NLI}-based Models for Inconsistency Detection in Summarization},
 url = {https://aclanthology.org/2022.tacl-1.10},
 volume = {10},
 year = {2022}
}

@inproceedings{DBLP:conf/nips/YuanNL21,
 author = {Weizhe Yuan and
Graham Neubig and
Pengfei Liu},
 bibsource = {dblp computer science bibliography, https://dblp.org},
 booktitle = {Advances in Neural Information Processing Systems 34: Annual Conference
on Neural Information Processing Systems 2021, NeurIPS 2021, December
6-14, 2021, virtual},
 editor = {Marc'Aurelio Ranzato and
Alina Beygelzimer and
Yann N. Dauphin and
Percy Liang and
Jennifer Wortman Vaughan},
 pages = {27263--27277},
 title = {{BARTS}core: Evaluating Generated Text as Text Generation},
 url = {https://proceedings.neurips.cc/paper/2021/hash/e4d2b6e6fdeca3e60e0f1a62fee3d9dd-Abstract.html},
 year = {2021}
}

@inproceedings{DBLP:conf/acl/TraininA25,
 address = {Vienna, Austria},
 author = {Trainin, Itamar  and
Abend, Omri},
 booktitle = {Findings of the Association for Computational Linguistics: ACL 2025},
 doi = {10.18653/v1/2025.findings-acl.1351},
 editor = {Che, Wanxiang  and
Nabende, Joyce  and
Shutova, Ekaterina  and
Pilehvar, Mohammad Taher},
 isbn = {979-8-89176-256-5},
 pages = {26347--26375},
 publisher = {Association for Computational Linguistics},
 title = {$T^5Score$: A Methodology for Automatically Assessing the Quality of {LLM} Generated Multi-Document Topic Sets},
 url = {https://aclanthology.org/2025.findings-acl.1351/},
 year = {2025}
}

@inproceedings{scire-etal-2024-fenice,
 address = {Bangkok, Thailand},
 author = {Scir{\`e}, Alessandro  and
Ghonim, Karim  and
Navigli, Roberto},
 booktitle = {Findings of the Association for Computational Linguistics: ACL 2024},
 doi = {10.18653/v1/2024.findings-acl.841},
 editor = {Ku, Lun-Wei  and
Martins, Andre  and
Srikumar, Vivek},
 pages = {14148--14161},
 publisher = {Association for Computational Linguistics},
 title = {{FENICE}: Factuality Evaluation of summarization based on Natural language Inference and Claim Extraction},
 url = {https://aclanthology.org/2024.findings-acl.841/},
 year = {2024}
}

@inproceedings{DBLP:conf/naacl/ZhongL25,
 author = {Yang Zhong and
Diane J. Litman},
 bibsource = {dblp computer science bibliography, https://dblp.org},
 booktitle = {Proceedings of the 2025 Conference of the Nations of the Americas
Chapter of the Association for Computational Linguistics: Human Language
Technologies, {NAACL} 2025 - Volume 1: Long Papers, Albuquerque, New
Mexico, USA, April 29 - May 4, 2025},
 doi = {10.18653/V1/2025.NAACL-LONG.103},
 editor = {Luis Chiruzzo and
Alan Ritter and
Lu Wang},
 pages = {2050--2073},
 publisher = {Association for Computational Linguistics},
 title = {Discourse-Driven Evaluation: Unveiling Factual Inconsistency in Long
Document Summarization},
 url = {https://doi.org/10.18653/v1/2025.naacl-long.103},
 year = {2025}
}

@inproceedings{lamsiyah-nourbakhsh-schommer:2025:RANLP,
    title = "Trust but Verify: A Comprehensive Survey of Faithfulness Evaluation Methods in Abstractive Text Summarization",
    author = "Lamsiyah, Salima  and
      Nourbakhsh, Aria  and
      Schommer, Christoph",
    editor = "Angelova, Galia  and
      Kunilovskaya, Maria  and
      Escribe, Marie  and
      Mitkov, Ruslan",
    booktitle = "Proceedings of the 15th International Conference on Recent Advances in Natural Language Processing - Natural Language Processing in the Generative AI Era",
    month = sep,
    year = "2025",
    address = "Varna, Bulgaria",
    publisher = "INCOMA Ltd., Shoumen, Bulgaria",
    url = "https://aclanthology.org/2025.ranlp-1.74/",
    pages = "633--643",
}

@inproceedings{DBLP:conf/naacl/HuangCPJW21,
 address = {Online},
 author = {Huang, Luyang  and
Cao, Shuyang  and
Parulian, Nikolaus  and
Ji, Heng  and
Wang, Lu},
 booktitle = {Proceedings of the 2021 Conference of the North American Chapter of the Association for Computational Linguistics: Human Language Technologies},
 doi = {10.18653/v1/2021.naacl-main.112},
 editor = {Toutanova, Kristina  and
Rumshisky, Anna  and
Zettlemoyer, Luke  and
Hakkani-Tur, Dilek  and
Beltagy, Iz  and
Bethard, Steven  and
Cotterell, Ryan  and
Chakraborty, Tanmoy  and
Zhou, Yichao},
 pages = {1419--1436},
 publisher = {Association for Computational Linguistics},
 title = {Efficient Attentions for Long Document Summarization},
 url = {https://aclanthology.org/2021.naacl-main.112},
 year = {2021}
}

@inproceedings{DBLP:conf/coling/YangW22,
 address = {Gyeongju, Republic of Korea},
 author = {Yang, Cai  and
Wan, Stephen},
 booktitle = {Proceedings of the Third Workshop on Scholarly Document Processing},
 editor = {Cohan, Arman  and
Feigenblat, Guy  and
Freitag, Dayne  and
Ghosal, Tirthankar  and
Herrmannova, Drahomira  and
Knoth, Petr  and
Lo, Kyle  and
Mayr, Philipp  and
Shmueli-Scheuer, Michal  and
de Waard, Anita  and
Wang, Lucy Lu},
 pages = {115--125},
 publisher = {Association for Computational Linguistics},
 title = {Investigating Metric Diversity for Evaluating Long Document Summarisation},
 url = {https://aclanthology.org/2022.sdp-1.13},
 year = {2022}
}

@inproceedings{DBLP:conf/naacl/GuoAUONSY22,
 address = {Seattle, United States},
 author = {Guo, Mandy  and
Ainslie, Joshua  and
Uthus, David  and
Ontanon, Santiago  and
Ni, Jianmo  and
Sung, Yun-Hsuan  and
Yang, Yinfei},
 booktitle = {Findings of the Association for Computational Linguistics: NAACL 2022},
 doi = {10.18653/v1/2022.findings-naacl.55},
 editor = {Carpuat, Marine  and
de Marneffe, Marie-Catherine  and
Meza Ruiz, Ivan Vladimir},
 pages = {724--736},
 publisher = {Association for Computational Linguistics},
 title = {{L}ong{T}5: {E}fficient Text-To-Text Transformer for Long Sequences},
 url = {https://aclanthology.org/2022.findings-naacl.55},
 year = {2022}
}

@inproceedings{DBLP:conf/aaai/SotudehCG21,
 author = {Sajad Sotudeh and
Arman Cohan and
Nazli Goharian},
 bibsource = {dblp computer science bibliography, https://dblp.org},
 booktitle = {Proceedings of the Workshop on Scientific Document Understanding co-located
with 35th {AAAI} Conference on Artificial Inteligence, SDU@AAAI 2021,
Virtual Event, February 9, 2021},
 editor = {Amir Pouran Ben Veyseh and
Franck Dernoncourt and
Thien Huu Nguyen and
Walter Chang and
Leo Anthony Celi},
 publisher = {CEUR-WS.org},
 series = {{CEUR} Workshop Proceedings},
 title = {On Generating Extended Summaries of Long Documents},
 url = {https://ceur-ws.org/Vol-2831/paper22.pdf},
 volume = {2831},
 year = {2021}
}

@inproceedings{yang-etal-2024-fizz,
 address = {Miami, Florida, USA},
 author = {Yang, Joonho  and
Yoon, Seunghyun  and
Kim, ByeongJeong  and
Lee, Hwanhee},
 booktitle = {Proceedings of the 2024 Conference on Empirical Methods in Natural Language Processing},
 doi = {10.18653/v1/2024.emnlp-main.3},
 editor = {Al-Onaizan, Yaser  and
Bansal, Mohit  and
Chen, Yun-Nung},
 pages = {30--45},
 publisher = {Association for Computational Linguistics},
 title = {{FIZZ}: Factual Inconsistency Detection by Zoom-in Summary and Zoom-out Document},
 url = {https://aclanthology.org/2024.emnlp-main.3/},
 year = {2024}
}

@inproceedings{DBLP:conf/acl/ZhaYLH23,
 address = {Toronto, Canada},
 author = {Zha, Yuheng  and
Yang, Yichi  and
Li, Ruichen  and
Hu, Zhiting},
 booktitle = {Proceedings of the 61st Annual Meeting of the Association for Computational Linguistics (Volume 1: Long Papers)},
 doi = {10.18653/v1/2023.acl-long.634},
 editor = {Rogers, Anna  and
Boyd-Graber, Jordan  and
Okazaki, Naoaki},
 pages = {11328--11348},
 publisher = {Association for Computational Linguistics},
 title = {{A}lign{S}core: Evaluating Factual Consistency with A Unified Alignment Function},
 url = {https://aclanthology.org/2023.acl-long.634},
 year = {2023}
}

@inproceedings{DBLP:conf/emnlp/TangLD24,
 address = {Miami, Florida, USA},
 author = {Tang, Liyan  and
Laban, Philippe  and
Durrett, Greg},
 booktitle = {Proceedings of the 2024 Conference on Empirical Methods in Natural Language Processing},
 doi = {10.18653/v1/2024.emnlp-main.499},
 editor = {Al-Onaizan, Yaser  and
Bansal, Mohit  and
Chen, Yun-Nung},
 pages = {8818--8847},
 publisher = {Association for Computational Linguistics},
 title = {{M}ini{C}heck: Efficient Fact-Checking of {LLM}s on Grounding Documents},
 url = {https://aclanthology.org/2024.emnlp-main.499/},
 year = {2024}
}

@inproceedings{DBLP:conf/emnlp/Zhong0YMJLZJH22,
 address = {Abu Dhabi, United Arab Emirates},
 author = {Zhong, Ming  and
Liu, Yang  and
Yin, Da  and
Mao, Yuning  and
Jiao, Yizhu  and
Liu, Pengfei  and
Zhu, Chenguang  and
Ji, Heng  and
Han, Jiawei},
 booktitle = {Proceedings of the 2022 Conference on Empirical Methods in Natural Language Processing},
 doi = {10.18653/v1/2022.emnlp-main.131},
 editor = {Goldberg, Yoav  and
Kozareva, Zornitsa  and
Zhang, Yue},
 pages = {2023--2038},
 publisher = {Association for Computational Linguistics},
 title = {Towards a Unified Multi-Dimensional Evaluator for Text Generation},
 url = {https://aclanthology.org/2022.emnlp-main.131},
 year = {2022}
}

@inproceedings{maynez-etal-2020-faithfulness,
 address = {Online},
 author = {Maynez, Joshua  and
Narayan, Shashi  and
Bohnet, Bernd  and
McDonald, Ryan},
 booktitle = {Proceedings of the 58th Annual Meeting of the Association for Computational Linguistics},
 doi = {10.18653/v1/2020.acl-main.173},
 editor = {Jurafsky, Dan  and
Chai, Joyce  and
Schluter, Natalie  and
Tetreault, Joel},
 pages = {1906--1919},
 publisher = {Association for Computational Linguistics},
 title = {On Faithfulness and Factuality in Abstractive Summarization},
 url = {https://aclanthology.org/2020.acl-main.173},
 year = {2020}
}

@inproceedings{wang-etal-2020-asking,
 address = {Online},
 author = {Wang, Alex  and
Cho, Kyunghyun  and
Lewis, Mike},
 booktitle = {Proceedings of the 58th Annual Meeting of the Association for Computational Linguistics},
 doi = {10.18653/v1/2020.acl-main.450},
 editor = {Jurafsky, Dan  and
Chai, Joyce  and
Schluter, Natalie  and
Tetreault, Joel},
 pages = {5008--5020},
 publisher = {Association for Computational Linguistics},
 title = {Asking and Answering Questions to Evaluate the Factual Consistency of Summaries},
 url = {https://aclanthology.org/2020.acl-main.450},
 year = {2020}
}

@inproceedings{scialom-etal-2021-questeval,
 address = {Online and Punta Cana, Dominican Republic},
 author = {Scialom, Thomas  and
Dray, Paul-Alexis  and
Lamprier, Sylvain  and
Piwowarski, Benjamin  and
Staiano, Jacopo  and
Wang, Alex  and
Gallinari, Patrick},
 booktitle = {Proceedings of the 2021 Conference on Empirical Methods in Natural Language Processing},
 doi = {10.18653/v1/2021.emnlp-main.529},
 editor = {Moens, Marie-Francine  and
Huang, Xuanjing  and
Specia, Lucia  and
Yih, Scott Wen-tau},
 pages = {6594--6604},
 publisher = {Association for Computational Linguistics},
 title = {{Q}uest{E}val: Summarization Asks for Fact-based Evaluation},
 url = {https://aclanthology.org/2021.emnlp-main.529},
 year = {2021}
}

@inproceedings{fabbri-etal-2022-qafacteval,
 address = {Seattle, United States},
 author = {Fabbri, Alexander  and
Wu, Chien-Sheng  and
Liu, Wenhao  and
Xiong, Caiming},
 booktitle = {Proceedings of the 2022 Conference of the North American Chapter of the Association for Computational Linguistics: Human Language Technologies},
 doi = {10.18653/v1/2022.naacl-main.187},
 editor = {Carpuat, Marine  and
de Marneffe, Marie-Catherine  and
Meza Ruiz, Ivan Vladimir},
 pages = {2587--2601},
 publisher = {Association for Computational Linguistics},
 title = {{QAF}act{E}val: Improved {QA}-Based Factual Consistency Evaluation for Summarization},
 url = {https://aclanthology.org/2022.naacl-main.187},
 year = {2022}
}

@article{chen-eger-2023-menli,
 address = {Cambridge, MA},
 author = {Chen, Yanran  and
Eger, Steffen},
 doi = {10.1162/tacl_a_00576},
 journal = {Transactions of the Association for Computational Linguistics},
 pages = {804--825},
 publisher = {MIT Press},
 title = {{MENLI}: Robust Evaluation Metrics from Natural Language Inference},
 url = {https://aclanthology.org/2023.tacl-1.47},
 volume = {11},
 year = {2023}
}

@inproceedings{fu-etal-2024-gptscore,
 address = {Mexico City, Mexico},
 author = {Fu, Jinlan  and
Ng, See-Kiong  and
Jiang, Zhengbao  and
Liu, Pengfei},
 booktitle = {Proceedings of the 2024 Conference of the North American Chapter of the Association for Computational Linguistics: Human Language Technologies (Volume 1: Long Papers)},
 editor = {Duh, Kevin  and
Gomez, Helena  and
Bethard, Steven},
 pages = {6556--6576},
 publisher = {Association for Computational Linguistics},
 title = {{GPTS}core: Evaluate as You Desire},
 url = {https://aclanthology.org/2024.naacl-long.365},
 year = {2024}
}

@inproceedings{wang-etal-2024-factcheck,
 address = {Miami, Florida, USA},
 author = {Wang, Yuxia  and
Gangi Reddy, Revanth  and
Mujahid, Zain Muhammad  and
Arora, Arnav  and
Rubashevskii, Aleksandr  and
Geng, Jiahui  and
Mohammed Afzal, Osama  and
Pan, Liangming  and
Borenstein, Nadav  and
Pillai, Aditya  and
Augenstein, Isabelle  and
Gurevych, Iryna  and
Nakov, Preslav},
 booktitle = {Findings of the Association for Computational Linguistics: EMNLP 2024},
 doi = {10.18653/v1/2024.findings-emnlp.830},
 editor = {Al-Onaizan, Yaser  and
Bansal, Mohit  and
Chen, Yun-Nung},
 pages = {14199--14230},
 publisher = {Association for Computational Linguistics},
 title = {Factcheck-{B}ench: Fine-Grained Evaluation Benchmark for Automatic Fact-checkers},
 url = {https://aclanthology.org/2024.findings-emnlp.830/},
 year = {2024}
}

@article{DBLP:journals/tois/HuangYMZFWCPFQL25,
 address = {New York, NY, USA},
 articleno = {42},
 author = {Huang, Lei and Yu, Weijiang and Ma, Weitao and Zhong, Weihong and Feng, Zhangyin and Wang, Haotian and Chen, Qianglong and Peng, Weihua and Feng, Xiaocheng and Qin, Bing and Liu, Ting},
 doi = {10.1145/3703155},
 issn = {1046-8188},
 issue_date = {March 2025},
 journal = {ACM Trans. Inf. Syst.},
 number = {2},
 numpages = {55},
 publisher = {Association for Computing Machinery},
 title = {A Survey on Hallucination in Large Language Models: Principles, Taxonomy, Challenges, and Open Questions},
 url = {https://doi.org/10.1145/3703155},
 volume = {43},
 year = {2025}
}

@inproceedings{belem-etal-2025-single,
 author = {Catarina G. Bel{\'{e}}m and
Pouya Pezeshkpour and
Hayate Iso and
Seiji Maekawa and
Nikita Bhutani and
Estevam Hruschka},
 bibsource = {dblp computer science bibliography, https://dblp.org},
 booktitle = {Findings of the Association for Computational Linguistics: {NAACL}
2025, Albuquerque, New Mexico, USA, April 29 - May 4, 2025},
 doi = {10.18653/V1/2025.FINDINGS-NAACL.293},
 editor = {Luis Chiruzzo and
Alan Ritter and
Lu Wang},
 pages = {5276--5309},
 publisher = {Association for Computational Linguistics},
 title = {From Single to Multi: How {LLMs} Hallucinate in Multi-Document Summarization},
 url = {https://doi.org/10.18653/v1/2025.findings-naacl.293},
 year = {2025}
}

@inproceedings{DBLP:conf/emnlp/LabanFXW24,
 address = {Miami, Florida, USA},
 author = {Laban, Philippe  and
Fabbri, Alexander  and
Xiong, Caiming  and
Wu, Chien-Sheng},
 booktitle = {Proceedings of the 2024 Conference on Empirical Methods in Natural Language Processing},
 doi = {10.18653/v1/2024.emnlp-main.552},
 editor = {Al-Onaizan, Yaser  and
Bansal, Mohit  and
Chen, Yun-Nung},
 pages = {9885--9903},
 publisher = {Association for Computational Linguistics},
 title = {Summary of a Haystack: A Challenge to Long-Context {LLM}s and {RAG} Systems},
 url = {https://aclanthology.org/2024.emnlp-main.552/},
 year = {2024}
}

@inproceedings{DBLP:conf/emnlp/GoldmanJSMDT24,
 address = {Miami, Florida, USA},
 author = {Goldman, Omer  and
Jacovi, Alon  and
Slobodkin, Aviv  and
Maimon, Aviya  and
Dagan, Ido  and
Tsarfaty, Reut},
 booktitle = {Proceedings of the 2024 Conference on Empirical Methods in Natural Language Processing},
 doi = {10.18653/v1/2024.emnlp-main.924},
 editor = {Al-Onaizan, Yaser  and
Bansal, Mohit  and
Chen, Yun-Nung},
 pages = {16576--16586},
 publisher = {Association for Computational Linguistics},
 title = {Is It Really Long Context if All You Need Is Retrieval? Towards Genuinely Difficult Long Context {NLP}},
 url = {https://aclanthology.org/2024.emnlp-main.924/},
 year = {2024}
}

@inproceedings{wang-etal-2022-squality,
 address = {Abu Dhabi, United Arab Emirates},
 author = {Wang, Alex  and
Pang, Richard Yuanzhe  and
Chen, Angelica  and
Phang, Jason  and
Bowman, Samuel R.},
 booktitle = {Proceedings of the 2022 Conference on Empirical Methods in Natural Language Processing},
 doi = {10.18653/v1/2022.emnlp-main.75},
 editor = {Goldberg, Yoav  and
Kozareva, Zornitsa  and
Zhang, Yue},
 pages = {1139--1156},
 publisher = {Association for Computational Linguistics},
 title = {{SQ}u{ALITY}: Building a Long-Document Summarization Dataset the Hard Way},
 url = {https://aclanthology.org/2022.emnlp-main.75},
 year = {2022}
}

@inproceedings{t-y-s-s-etal-2024-lexabsumm,
 address = {Torino, Italia},
 author = {T.y.s.s., Santosh  and
Aly, Mahmoud  and
Grabmair, Matthias},
 booktitle = {Proceedings of the 2024 Joint International Conference on Computational Linguistics, Language Resources and Evaluation (LREC-COLING 2024)},
 editor = {Calzolari, Nicoletta  and
Kan, Min-Yen  and
Hoste, Veronique  and
Lenci, Alessandro  and
Sakti, Sakriani  and
Xue, Nianwen},
 pages = {10422--10431},
 publisher = {ELRA and ICCL},
 title = {{L}ex{A}b{S}umm: Aspect-based Summarization of Legal Decisions},
 url = {https://aclanthology.org/2024.lrec-main.911},
 year = {2024}
}

@inproceedings{DBLP:conf/emnlp/QiuZKPC23,
 address = {Singapore},
 author = {Qiu, Yifu  and
Ziser, Yftah  and
Korhonen, Anna  and
Ponti, Edoardo  and
Cohen, Shay},
 booktitle = {Proceedings of the 2023 Conference on Empirical Methods in Natural Language Processing},
 doi = {10.18653/v1/2023.emnlp-main.551},
 editor = {Bouamor, Houda  and
Pino, Juan  and
Bali, Kalika},
 pages = {8914--8932},
 publisher = {Association for Computational Linguistics},
 title = {Detecting and Mitigating Hallucinations in Multilingual Summarisation},
 url = {https://aclanthology.org/2023.emnlp-main.551},
 year = {2023}
}

@inproceedings{DBLP:conf/acl/ZhangPZW24,
 author = {Shuo Zhang and
Liangming Pan and
Junzhou Zhao and
William Yang Wang},
 bibsource = {dblp computer science bibliography, https://dblp.org},
 booktitle = {Findings of the Association for Computational Linguistics, {ACL} 2024,
Bangkok, Thailand and virtual meeting, August 11-16, 2024},
 doi = {10.18653/V1/2024.FINDINGS-ACL.121},
 editor = {Lun{-}Wei Ku and
Andre Martins and
Vivek Srikumar},
 pages = {2025--2038},
 publisher = {Association for Computational Linguistics},
 title = {The Knowledge Alignment Problem: Bridging Human and External Knowledge
for Large Language Models},
 url = {https://doi.org/10.18653/v1/2024.findings-acl.121},
 year = {2024}
}

@inproceedings{DBLP:conf/iclr/MundlerHJV24,
 author = {Niels M{\"{u}}ndler and
Jingxuan He and
Slobodan Jenko and
Martin T. Vechev},
 bibsource = {dblp computer science bibliography, https://dblp.org},
 booktitle = {The Twelfth International Conference on Learning Representations,
{ICLR} 2024, Vienna, Austria, May 7-11, 2024},
 publisher = {OpenReview.net},
 title = {Self-contradictory Hallucinations of Large Language Models: Evaluation,
Detection and Mitigation},
 url = {https://openreview.net/forum?id=EmQSOi1X2f},
 year = {2024}
}

@inproceedings{DBLP:conf/acl/GaoDPCCFZLLJG23,
 address = {Toronto, Canada},
 author = {Gao, Luyu  and
Dai, Zhuyun  and
Pasupat, Panupong  and
Chen, Anthony  and
Chaganty, Arun Tejasvi  and
Fan, Yicheng  and
Zhao, Vincent  and
Lao, Ni  and
Lee, Hongrae  and
Juan, Da-Cheng  and
Guu, Kelvin},
 booktitle = {Proceedings of the 61st Annual Meeting of the Association for Computational Linguistics (Volume 1: Long Papers)},
 doi = {10.18653/v1/2023.acl-long.910},
 editor = {Rogers, Anna  and
Boyd-Graber, Jordan  and
Okazaki, Naoaki},
 pages = {16477--16508},
 publisher = {Association for Computational Linguistics},
 title = {{RARR}: Researching and Revising What Language Models Say, Using Language Models},
 url = {https://aclanthology.org/2023.acl-long.910},
 year = {2023}
}

@article{DBLP:journals/tacl/LiuLHPBPL24,
 address = {Cambridge, MA},
 author = {Liu, Nelson F.  and
Lin, Kevin  and
Hewitt, John  and
Paranjape, Ashwin  and
Bevilacqua, Michele  and
Petroni, Fabio  and
Liang, Percy},
 doi = {10.1162/tacl_a_00638},
 journal = {Transactions of the Association for Computational Linguistics},
 pages = {157--173},
 publisher = {MIT Press},
 title = {Lost in the Middle: How Language Models Use Long Contexts},
 url = {https://aclanthology.org/2024.tacl-1.9},
 volume = {12},
 year = {2024}
}

@inproceedings{DBLP:conf/acl/Bird06,
 address = {Philadelphia, Pennsylvania, USA},
 author = {Loper, Edward  and
Bird, Steven},
 booktitle = {Proceedings of the {ACL}-02 Workshop on Effective Tools and Methodologies for Teaching Natural Language Processing and Computational Linguistics},
 doi = {10.3115/1118108.1118117},
 pages = {63--70},
 publisher = {Association for Computational Linguistics},
 title = {{NLTK}: The Natural Language Toolkit},
 url = {https://aclanthology.org/W02-0109},
 year = {2002}
}

@inproceedings{reimers-gurevych-2019-sentence,
 address = {Hong Kong, China},
 author = {Reimers, Nils  and
Gurevych, Iryna},
 booktitle = {Proceedings of the 2019 Conference on Empirical Methods in Natural Language Processing and the 9th International Joint Conference on Natural Language Processing (EMNLP-IJCNLP)},
 doi = {10.18653/v1/D19-1410},
 editor = {Inui, Kentaro  and
Jiang, Jing  and
Ng, Vincent  and
Wan, Xiaojun},
 pages = {3982--3992},
 publisher = {Association for Computational Linguistics},
 title = {Sentence-{BERT}: Sentence Embeddings using {S}iamese {BERT}-Networks},
 url = {https://aclanthology.org/D19-1410},
 year = {2019}
}
